\documentclass{article} 

 \usepackage[table]{xcolor} 

\usepackage[final]{neurips_2023}

\usepackage{graphicx}
\usepackage{float}

\usepackage{mathptmx}

\usepackage{latexsym}
\usepackage{tcolorbox}

\usepackage{wrapfig,lipsum,booktabs}

\usepackage{url}            
\usepackage{amsfonts}       
\usepackage{nicefrac}       
\usepackage{multirow}

\usepackage{amsmath}
\usepackage{amssymb}
\usepackage{mathtools}
\usepackage{amsthm}
\usepackage{balance}
\usepackage{flushend}
\usepackage{makecell}

\usepackage{listings}   
\usepackage{algorithm}      
\usepackage{algpseudocode}  

\usepackage{lipsum}

\usepackage{tabularx}

\usepackage{pifont}  

\usepackage{enumitem}
\usepackage{booktabs,arydshln}

\usepackage{hyperref}        
\usepackage{cleveref}
\definecolor{lightgreen}{rgb}{0.828,0.914,0.910}
\definecolor{airforceblue}{rgb}{0.08, 0.38, 0.74}
\newcolumntype{g}{>{\columncolor{airforceblue}}c}
\newcommand{\cmark}{\ding{51}} 
\newcommand{\xmark}{\ding{55}} 
\definecolor{mgreen2}{rgb}{0.2, 0.521, 0.502}
\hypersetup{
	colorlinks,
	citecolor=darkgray,
	linkcolor=black,
	urlcolor=blue}

\newcommand{\sysname}{{rStar2-Agent}}

\title{rStar2-Agent: Agentic Reasoning Technical Report}

\author{Ning Shang$^{*}$\hspace{5pt} Yifei Liu$^{*}$\hspace{5pt}   Yi Zhu$^{*}$\hspace{5pt}  Li Lyna Zhang$^{*\dagger}$\hspace{3pt} \\ \bf Weijiang Xu\hspace{3pt} Xinyu Guan\hspace{3pt} Buze Zhang\hspace{3pt} Bingcheng Dong\hspace{3pt}  Xudong Zhou\hspace{3pt} Bowen Zhang\hspace{3pt} \\\bf Ying Xin \hspace{5pt} Ziming Miao \hspace{5pt} Scarlett Li \hspace{5pt}  Fan Yang \hspace{5pt} Mao Yang$^{\dagger}$
	\\\\\fontsize{11}{11} \selectfont{Microsoft Research}  	
}

\begin{document}
	\newcommand{\todo}[1]{\textcolor{red}{TODO: #1}}
		\newcommand{\lz}[1]{\textcolor{blue}{Lyna: #1}}
	\maketitle

\def\thefootnote{$*$}\footnotetext{The first four authors contributed equally}
\def\thefootnote{$\dagger$}\footnotetext{Project leaders; correspondence to lzhani@microsoft.com and maoyang@microsoft.com}

\vspace{-3ex}
\begin{abstract}
	\vspace{-2ex}
	We introduce {\sysname},  a 14B math reasoning model trained with agentic reinforcement learning to achieve frontier-level performance. Beyond current long CoT, the model demonstrates advanced cognitive behaviors, such as  thinking carefully before using Python coding tools and  reflecting on code execution feedback to autonomously explore, verify, and refine intermediate steps in complex problem-solving. This capability is enabled through three key innovations that makes agentic RL effective at scale:  \textbf{(i)} an efficient RL infrastructure with a reliable Python code environment that supports high-throughput  execution and mitigates the high rollout costs, enabling  training on limited GPU resources (64 MI300X GPUs); \textbf{(ii)} \textit{GRPO-RoC}, an agentic RL algorithm with a \textit{Resample-on-Correct} rollout strategy that addresses the inherent environment noises from coding tools, allowing the model to reason more effectively  in a  code environment; \textbf{(iii)} An efficient agent training recipe that starts with \textit{non-reasoning} SFT and progresses through multi-RL stages, yielding advanced cognitive abilities with minimal compute cost.
 To this end, {\sysname} boosts a pre-trained 14B model to state of the art  in only 510 RL steps within one week, achieving average pass@1 scores of 80.6\% on AIME24 and 69.8\% on AIME25, surpassing DeepSeek-R1 (671B) with significantly shorter responses.  Beyond mathematics, {\sysname}-14B also demonstrates strong generalization to alignment, scientific reasoning, and agentic tool-use tasks. Code and training recipes are available at \url{https://github.com/microsoft/rStar}.
	
	\begin{table}[h]
		\centering
		\vspace{-1ex}
		\label{tbl:teaser}
		\begin{tabular}{cccc}
			\toprule
			{Model}	     & AIME24&AIME25&HMMT25\\
			\midrule
			OpenAI o3-mini (medium) & 79.6 & \bf77.0 & \bf53.0\\
			DeepSeek-R1 (671B) & 79.8 & 70.0 & 44.4\\
			DeepSeek-R1-Zero (671B) &71.0 & 53.3&46.0 \\
			Claude-Opus-4.0 (Think) & 76.0 & 69.2 & -\\
			QWQ-32B & 79.5& 65.8& 47.5\\
			\rowcolor{lightgreen}	\bf\sysname-14B & \bf 80.6 & 69.8 & 52.7\\
			\hline
		\end{tabular}%
	\vspace{-3ex}
\end{table}

\begin{figure}[h]
	\centering
	\includegraphics[width=0.75\textwidth]{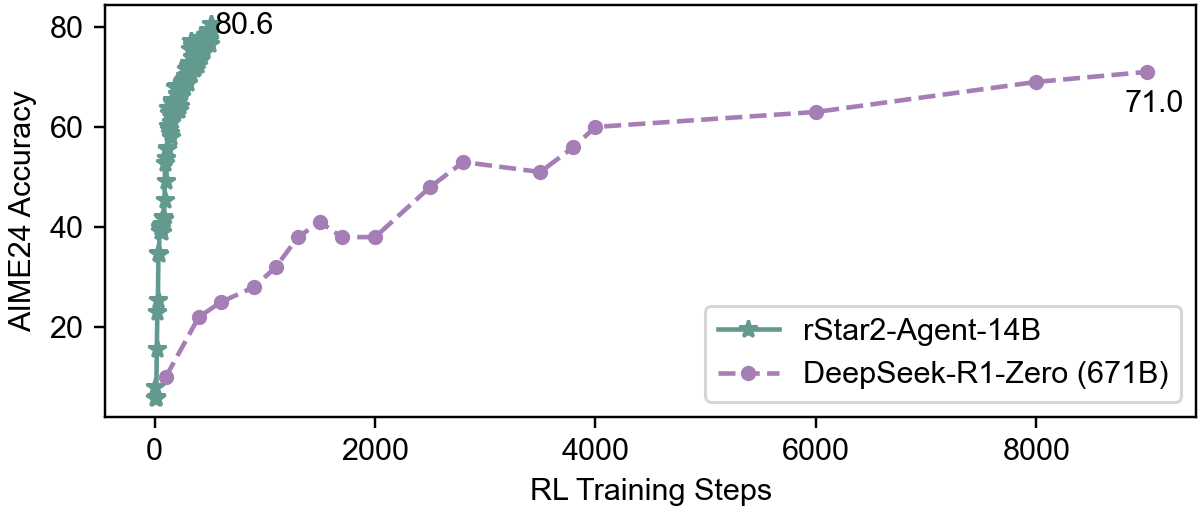}
	\vspace{-1ex}	
	\caption{{\sysname}-14B reaches frontier-level math reasoning in just 510 RL training steps.}	
	\label{fig:teaser}
\end{figure}

\end{abstract}
\tableofcontents 
\newpage
\vspace{-2ex}
\section{Introduction}
\vspace{-2ex}

Test-time scaling has recently driven substantial advances in complex reasoning~\citep{rstarmath,kimi1.5}. Leading models such as OpenAI o-series~\citep{o1,o1_reason},  DeepSeek-R1~\citep{r1} and Gemini-2.5~\citep{gemini}  show that extending the Chain-of-Thought (CoT), in essence ``\emph{thinking longer}", can markedly improve performance, especially when optimized through large-scale reinforcement learning with verifiable rewards (RLVR).  However, long CoT remains fundamentally limited for hard problems prone to subtle intermediate errors or requiring creative shifts in reasoning. 
In these cases, models depend on internal self-reflection, which often fails to detect mistakes~\citep{sui2025stop} or to self-correct when the initial approach is flawed. 

To move beyond merely “thinking longer", we aim to enable models to “\textbf{think smarter}" by developing more advanced cognitive abilities that autonomously utilize the right tools to reason, validate, and learn from the feedback signals provided by the tool environment.   We incentivize these abilities through \textit{agentic reinforcement learning}, where the model interacts with tools insides the dedicated tool environment and adapts its reasoning based on the feedback it receives. Crucially, not all tools or environments are equally effective; a valuable environment must be deployable and provide accurate, verifiable signals that guide the model toward stronger reasoning paths. In this work, we focus on \textit{Python coding tools and the interpreter} as the environment for agentic reinforcement learning.    Python coding tools broaden the model's action space, enabling  exploration of alternative solutions and verification of intermediate steps, thereby complementing internal self-reflection when long CoT alone is insufficient.

However, effectively scaling agentic reinforcement learning poses significant challenges. First, the inherent complexity of coding tools and Python interpreter introduces environment noise into the reasoning process. When the model inevitably generates syntactically or logically incorrect code, the resulting environment feedback (e.g., error message) can cause  it to waste valuable tokens correcting mistakes rather than advancing reasoning. Unfortunately, current RL methods~\citep{shao2024deepseekmath,r1}, which rely primarily on outcome-only rewards,   exacerbates this issue because trajectories with failed intermediate tool calls still receive positive reward if the final answer is correct. As a result,  the model treats errors as acceptable and produces lengthy, low-quality reasoning trajectory. 
Second, large-scale agentic RL training imposes substantial infrastructure demands.   A single training batch can trigger tens of thousands of concurrent tool calls, making it challenging to construct a reliable and responsive code execution environment. Moreover,  agentic rollouts with environment interactions amplify the rollout inefficiencies in standard RL systems, significantly slowing the overall training process.

In this work, we introduce \textbf{\sysname}, a novel agentic reinforcement learning approach that trains a 14B reasoning model, {\sysname}-14B, to reach frontier-level performance, rivaling or surpassing the 671B DeepSeek-R1. 
 {\sysname} incorporates three key innovations. First, we build an efficient and reliable infrastructure for large-scale agentic RL.   We construct  a high-throughput, isolated code environment  capable of handling   $~$45K concurrent tool calls, with execution  feedback returned in just 0.3 seconds on average.    To address RL rollout inefficiencies, we introduce a load-balanced rollout scheduler that dynamically allocates rollout requests based on available KV cache capacity across GPUs to maximize  computational utilization. This infrastructure enables efficient RL training even with limited GPU resources. Using 64$\times$MI300X GPUs, we complete rStar2-Agent-14B training in just one week.

Second, to enable effective  agentic reinforcement learning in a code environment, we propose \emph{Group Relative Policy Optimization with Resampling on Correct (GRPO-RoC)}, which integrates GRPO with a \emph{Resample-On-Correct (RoC)} rollout strategy to  address environment-induced noise  under sparse, outcome-only rewards. Specifically, RoC first oversamples a larger group of rollouts and then downsamples to the standard batch size. Positive trajectories are filtered to retain only the highest-quality ones with minimal tool-induced errors or formatting issues, while negative trajectories are uniformly downsampled. This simple yet effective asymmetric sampling preserves diverse failure modes as informative negative signals while emphasizing higher-quality success cases for positive supervision. 
Compared to methods that explicitly penalize tool-use errors in the reward  function~\citep{toolrl,torl,kimiresearcher}, GRPO-RoC improves training stability and avoids reward-hacking risks. By learning from cleaner, higher-quality positive trajectories, the model not only improves Python coding tool usage but also exhibits advanced cognitive abilities, reasoning more effectively and concisely under realistic code-environment interactions.

Finally, we present our training recipe that boosts a 14B pre-trained base model to frontier-level math reasoning with minimal compute. Unlike prior works that apply reasoning-heavy SFT before RL~\citep{liu2025prorl,retool,qwen3technicalreport,seed2025seed1}, we begin with a \emph{non-reasoning} SFT stage solely to instill general instruction-following, coding tool usage, and formatting, without enhancing reasoning. This avoids potential SFT overfitting and keeps  initial average responses short, allowing RL to more effectively cultivate reasoning  while fully exploiting the model's pre-trained capability. We then conduct multi-stage RL training with GRPO-RoC, gradually  increasing
task difficulty and maximum training length. Unlike prior RL methods that heavily scale rollouts to 16K$\rightarrow$ 48K or more~\citep{chen2025minimax,xiaomi2025mimo}, we limit each stage to shorter lengths (8K$\rightarrow$12K). This significantly reduces RL costs while encouraging more efficient reasoning strategies. 
With only 510 RL steps, the model rapidly achieves frontier-level math reasoning, demonstrating both high capability and exceptional training efficiency.

The final resulting model, {\sysname}-14B, achieves strong math reasoning performance, surpassing leading reasoning models such as DeepSeek-R1 and Kimi k1.5. Remarkably, on AIME24, it reaches 80.6\% accuracy, outperforming  o3-mini (medium), DeepSeek-R1, and Claude Opus 4.0 (thinking) by 1.0\%, 0.8\% and 3.6\%, respectively, and reaches 69.8\% and 52.7\% on AIME25 and HMMT25, demonstrating consistently strong results. 
Beyond mathematics,  it generalizes effectively despite being trained with math-only agentic reinforcement learning. It outperforms  DeepSeek-V3 on the GPQA-Diamond science reasoning benchmark, excels at agentic tool use on BFCL v3, and delivers  competitive results on general benchmarks such as IFEval and Arena-Hard. We also report our unsuccessful attempts and analyses, highlighting the discovery of more advanced cognitive reasoning behaviors incentivized by {\sysname} agentic RL, such as reflection tokens on environment feedback that drive more effective reasoning.

\section{Agentic Reinforcement Learning Methodology}

\begin{figure*}[h]
	\centering
	\includegraphics[width=1\textwidth]{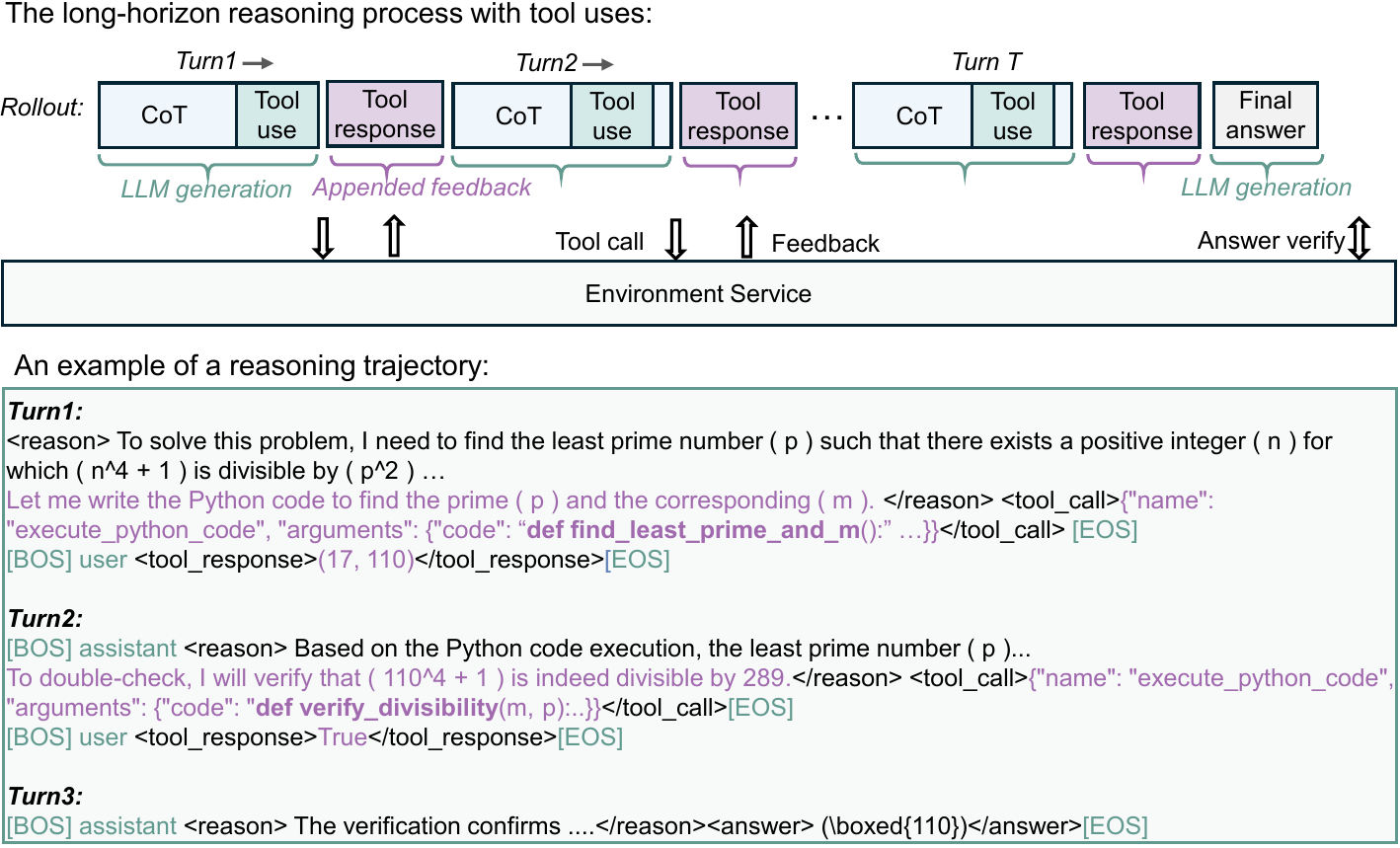}	
	\vspace{-4ex}
	\caption{{\sysname} trains LLMs to natively use Python coding tools within the dedicated execution environment,   enabling  more advanced and effective reasoning for complex problem-solving. }	
	\label{fig:cotexample}
\end{figure*}

\subsection{Smarter Reasoning in a Code Environment}
\label{sec:prompt}
Python code and its interpreter, along with scientific computing libraries such as \texttt{Numpy} for efficient numerical computation, \texttt{Scipy} for advanced scientific analysis, and \texttt{SymPy} for symbolic mathematics, can significantly improve the model's ability for math problem-solving. 
Ideally, the model demonstrates human-like cognitive behaviors in this Python code environment: \textit{(i)} invoking tools at the right reasoning steps; \textit{(ii)} writing logically correct and functional code, and \textit{(iii)} carefully  reflecting on  execution results to guide subsequent reasoning steps. 
We cultivate this capability through agentic reinforcement learning, and in this section, we  introduce our key design choices, including tool call interfaces and prompt templates.

\begin{figure}[t]
	\centering
	\includegraphics[width=1\textwidth]{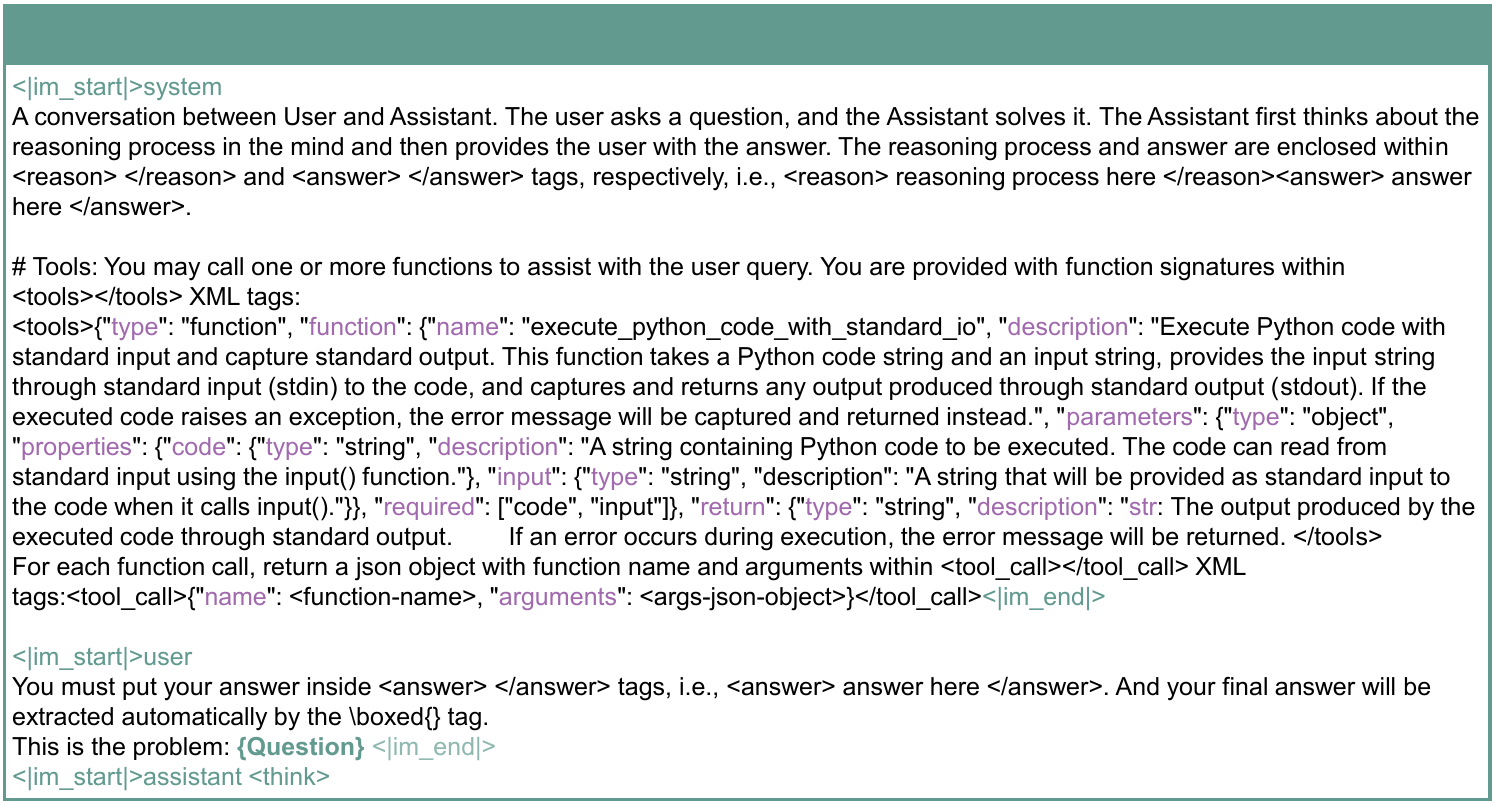}	
	\caption{Our prompt template. Question will be replaced with the specific  question during training.}	
	\label{fig:prompt}
\end{figure}

\noindent\textbf{Multi-turn Rollout}. With  coding tools, the model performs multi-turn rollouts that incorporate execution results  from the code environment into reasoning, as illustrated in Fig.~\ref{fig:cotexample}. Unlike standard RL rollouts, which generate a full trajectory until an EOS token,  we produce  full trajectories  through multiple interactive turns with the code environment. Specifically, the first turn begins with a predefined system prompt (Fig.~\ref{fig:prompt}) and the given question. Then the model generates an initial reasoning trajectory in the role of \texttt{assistant}, ending at the \texttt{EOS} token. If no code tool call is present, the rollout terminates. Otherwise, the code block is extracted and executed by the environment service, and the output is appended to the trajectory under the  \texttt{user} role. The model then takes this updated context as input and continues the next turn of reasoning under the \texttt{assistant} role. This multi-turn rollout process repeats until the model produces a final answer or reaches a predefined maximum number of turns $T$.

\noindent\textbf{Tool Call Format}.  We use a general function call interface for invoking coding tools, with each tool call represented in a structured JSON format as shown in the example below:

\begin{center}
	\texttt{$<$tool\_call$>$}\{``name": ``\textcolor{mgreen2}{execute\_python\_code\_with\_standard\_io}", ``arguments": \{``code": ``\emph{\textcolor{mgreen2}{import sympy\textbackslash n\textbackslash n    def verify\_divisibility (m,p):\textbackslash n\textbackslash n \quad  for p in sympy.primerange(2, 100000) $\cdots$} }              ", ``input": ``"\}\texttt{$<$/tool\_call$>$}
\end{center}
At the end of each turn, we check for \texttt{$<$tool\_call$>$} \texttt{$<$/tool\_call$>$} blocks. If found, the JSON is parsed to extract the code block from the ``\textcolor{mgreen2}{\texttt{code}}"  field and, if available, input arguments from the ``\textcolor{mgreen2}{\texttt{input}}" field within ``\textcolor{mgreen2}{\texttt{arguments}}". If parsing fails due to an invalid format, the error message is wrapped in \texttt{$<$tool\_response$>$}\texttt{$<$/tool\_response$>$} tags and returned to the model. Otherwise, the extracted code and arguments are forwarded to the environment service (see Fig.~\ref{fig:cotexample}), which produces one of four possible responses: 

\begin{itemize}
	\item \textit{successful execution with standard output}, returning the program output;
	\item \textit{successful execution without standard output}, returning the output as shown by IPython;
	\item \textit{execution error}, returning the error message and traceback logs; 
	\item \textit{timeout}, where the code is syntactically valid but fails to complete within the time limit, often due to high complexity or logical errors such as infinite loops. 
\end{itemize}
In all cases, the environment feedback is wrapped in \texttt{$<$tool\_response$>$} tags and fed back to the model.

This structured approach provides a standardized, API-like interface that removes parsing ambiguity and clearly separates reasoning from execution. Compared to previous methods~\citep{agentrl,torl,retool} that rely on markdown-style syntax (e.g., $^{\cdots}$python ... $^{\cdots}$and $^{\cdots}$output ...$^{\cdots}$) or custom tokens (e.g., $<$code$>$, $<$interpreter$>$), our design is more extensible, generalizes to diverse tools, and aligns with the widely-used function-calling protocols in LLM APIs, which facilitates integration and future extension.

\noindent\textbf{Prompt Template}. Fig.~\ref{fig:prompt} shows our prompt used during the agentic reinforcement learning. The model is instructed to first generate a reasoning process enclosed in  $<$\texttt{reason}$>\cdots$$</$\texttt{reason}$>$, followed by the final answer in $<$\texttt{answer}$>\cdots$$</$\texttt{answer}$>$.
To guide correct coding tool usage, the prompt explicitly details the available tools (i.e., the coding tool), including the descriptions and the structured function call format.

Notably, the multi-turn rollout setup may produce multiple  $<$\texttt{reason}$>$ blocks, but only a single $<$\texttt{answer}$>$ block is allowed, as shown in  Fig.~\ref{fig:cotexample}.  The final numeric result must be wrapped in  \textbackslash\texttt{boxed}\{\} for extraction.

\subsection{End-to-End Agentic Reinforcement Learning}
\subsubsection{Preliminary: GRPO}
\label{fig:test}

\noindent\textbf{Group Relative Policy Optimization (GRPO)}. We start by introducing the GRPO algorithm. Specifically, 
for each question $q$ and its ground-truth answer $a$ from a dataset ${D}$, GRPO samples a group of rollout trajectories $\{o_1,o_2,\cdots,o_{G}\}$ from the old policy $\pi_{\theta_{old}}$ and then optimizes the policy  $\pi_{\theta}$ by maximizing the following objective:

\begin{align}
	J_{\text{GRPO}}(\theta)
	&= \mathbb{E}_{(q,a) \sim {D},\, \{o_i\}_{i=1}^G \sim \pi_{\theta_{\text{old}}}(\cdot|q)} \notag \\
	&	\left[
	\frac{1}{G}\sum_{i=1}^{G}\frac{1}{|o_i|}\sum_{t}^{|o_i|} \left( \min\left[\frac{\pi_\theta(o_{i,t}|q,o_{i,<t})}{\pi_{\theta_{old}}(o_{i,t}|q,o_{i,<t})} {A}_{i,t},\; \mathrm{clip}(\frac{\pi_\theta(o_{i,t}|q,o_{i,<t})}{\pi_{\theta_{old}}(o_{i,t}|q,o_{i,<t})}, 1 - \varepsilon, 1 + \varepsilon)\, {A}_{i,t} \right] 
	- \beta D_{KL}(\pi_\theta\,\|\,\pi_{\text{ref}})\right)\right]
	\label{eq:grpo_loss}
\end{align}
where $\varepsilon$ and $\beta$ are hyper-parameters that control the clipping range of importance sampling ratio and the weight of KL penalty term, respectively. 
$A_{i,t}$ denotes the  estimated  advantage, computed using a group of rewards $\{r_1,r_2,...r_G\}$ corresponding to the outputs within each group: 
\begin{align}
A_{i,t}=\frac{r_i-\text{mean}(\{r_1,r_2,\cdots,r_G\})}{\text{std}(\{r_1,r_2,\cdots,r_G\})}
\end{align}
Here, $r_i$ is the reward assigned to rollout trajectory $o_i$, which is evaluated via a rule-based verifier system to mitigate  reward hacking~\citep{r1,kimi1.5}.

\noindent\textbf{Outcome-only Reward Design}. Recent RL methods for math reasoning have seen substantial gains by using outcome-only rewards, a key design choice that effectively avoids reward hacking~\citep{r1,kimi1.5}.
Specifically, each rollout trajectory $o_i$ receives a binary accuracy reward $r_i\in \{0,1\}$ based on whether the final answer matches the ground truth answer $a$: 
\begin{align}
	\label{eq:reward}
	\quad r_{i}=\begin{cases} 1 &\text{if \texttt{is\_equivalent}($a$, $o_i$)}, \\ 0 & \text{otherwise}. \end{cases}
\end{align}
In math word problems, we extract the final answer from \textbackslash\texttt{boxed}\{\}  within the $<$\texttt{answer}$>$ tag and verify it against the ground truth $a$ using the rule-based \texttt{math\_verify} tool.   Correct matches get a reward of 1, mismatches receive 0.

\noindent\textbf{More Exploration}. To push the policy beyond its pre-training limits, we incorporate several key modifications from recent works.  First, we remove the KL divergence penalty. Although commonly used to prevent the online policy from significantly deviating from a reference policy and to stabilize training, it can inadvertently restrict the discovery of novel, tool-augmented reasoning patterns.  Removing it allows the model to explore more freely. 

Second, we adopt the \textit{Clip-Higher}~\citep{dapo} strategy by relaxing the upper bound of the importance sampling ratio.  Specifically, we follow prior work and increase $\varepsilon_\text{high}$ from 0.2 to 0.28, allowing the model to better explore high-entropy, low-probability tokens. These minority tokens may include forking tokens that are essential for reasoning performance, as noted in recent studies~\citep{wang2025beyond,cheng2025reasoning}.

Third, we eliminate the entropy loss term to prevent training instability. While commonly used to encourage exploration,  it can cause uncontrolled entropy growth,  potentially leading to training collapse.

\subsubsection{Challenges in Agentic Reinforcement Learning}
\label{sec:toolerros}
\begin{figure}[t]
\centering
\includegraphics[width=1\textwidth]{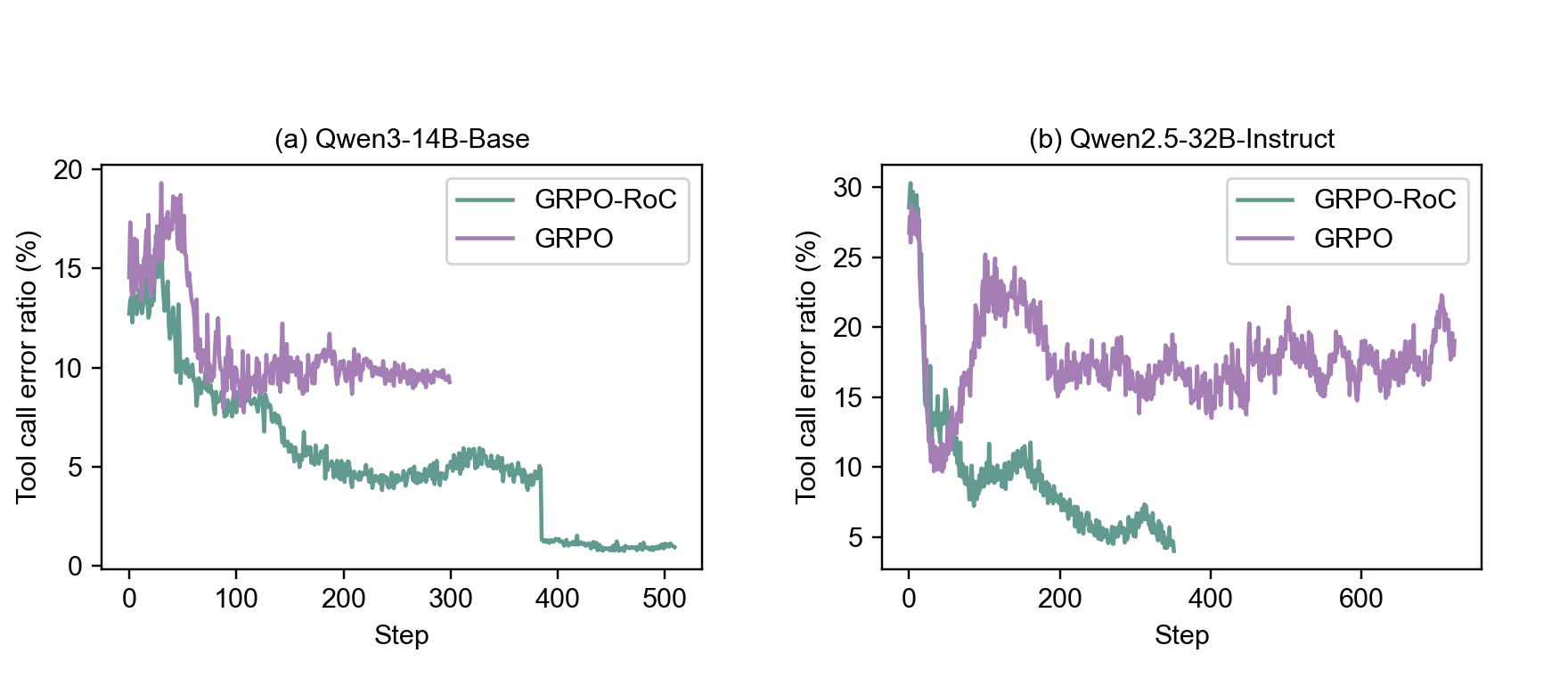}	
\vspace{-4ex}
\caption{Proportion of tool calls that contain errors within correctly answered trajectories. Under naive GRPO, the error rate initially decreases but soon plateaus at a significant level. In contrast, our GRPO-RoC continues to reduce tool-related errors with more training steps.}	
\label{fig:tcerror}
\end{figure}

\noindent\textbf{Inherent Environment Noises}. While GRPO provides a strong foundation, agentic reinforcement learning introduces new challenges. In particular, coding tools and the code environment introduce inherent noise into reasoning. 

Unlike standard reasoning, coding tools require the model not only to decide when to use them but also to generate correct and executable code for the intended functionality. When  errors occur, the  environment returns error messages unrelated to the reasoning task. This \textbf{noisy feedback} can mislead the model, causing it to spend valuable tokens fixing tool errors rather than advancing its reasoning. Such distractions significantly hinder problem-solving, whereas they do not occur in pure CoT reasoning.

\noindent\textbf{Impact of Outcome-only Reward on Trajectory Quality}. Under current outcome-only reward schemes, trajectories are evaluated solely based on the final answer to prevent reward hacking. However, this outcome-only reward cannot penalize undesirable intermediate behaviors. 
As a result, trajectories with incorrect intermediate tool calls can still receive positive reward if the final answer is correct, effectively reinforcing the model to treat such errors are acceptable. As shown in Fig.~\ref{fig:tcerror}, under naive GRPO, the ratio of tool-related errors in \textit{positively rewarded} trajectories initially decreases but eventually stabilizes at a significant level, around 15\% for Qwen2.5-32B and 10\% for Qwen3-14B. Consequently,  the model tends to produce lengthy, low-quality trajectories containing tool call errors, severely limiting the effectiveness of agentic reinforcement learning and inflating training costs.

\subsubsection{GRPO-RoC: Group Relative Policy Optimization with Resampling on Correct}
For more effective agentic reinforcement learning, we introduce Group Relative Policy Optimization with Resampling on Correct (GRPO-RoC). This section details our design choice and methodology.

\noindent\textbf{Design Principle: Answer-only Outcome Reward}. Environment noise can cause the model to generate lengthy, low-quality  but correctly answered trajectories.   From a reward design perspective, two potential solutions exist:
 \textit{(i)} introducing step-level reward~\citep{yue2025promoting}; \textit{(ii)} retaining outcome-only rewards while adding penalties, such as for tool-call errors~\citep{toolrl,torl,kimiresearcher}.  However, we do not adopt these approaches for two main reasons: \textbf{(i)} they introduce additional complexity, such as requiring careful human tuning and  reward model construction; \textbf{(ii)} they are prone to reward hacking. For example, during early training, when the model's reasoning ability is still developing,  step-level rewards or  tool-error penalties can hinder effective exploration.

 To avoid reward hacking, we use a minimal  answer-only outcome reward, as shown in Eq.~\ref{eq:reward}. To address the challenge introduced by environment noise, we introduce GRPO-RoC, which effectively filters out low-quality noisy trajectories  through  \textit{Resample on Correct (RoC)} rollout strategy.

\noindent\textbf{Resample on Correct (RoC)} is a simple yet effective rollout strategy that enables effective agentic reinforcement learning under an answer-only outcome reward regime. Specifically,  we first oversample a larger group of rollouts and then downsample to the standard rollout batch size. Positive trajectories are filtered to retain only the highest-quality ones with minimal tool-induced errors or tool call formatting issues, while negative trajectories are uniformly downsampled. This asymmetric sampling reinforces positive supervision without losing the various learning signal from failures, facilitating more effective policy updates.  Although generally applicable to various RL algorithms, in this work, we instantiate RoC on GRPO, resulting in the algorithm \textbf{GRPO-RoC}.

In standard GRPO, each question is sampled with a group of $G$ rollout trajectories $\{o_i\}^G_{i=1}$, which are then used to compute rewards and update the policy. In our GRPO-RoC, we first oversample \textcolor{mgreen2}{$2G$ rollouts  trajectories $\{o_i\}^{2G}_{i=1}$} and then apply the RoC strategy to select $G$ trajectories for policy updates. Specifically,  let $O_\text{neg}=\{o_i^\text{neg}\}$ and $O_\text{pos}=\{o_i^\text{pos}\}$  denote the group of negatively and positively rewarded trajectories, respectively, where $|O_\text{neg}| + |O_\text{pos}| = 2G$. We then apply different selection strategies to each group: we sample $\hat{O}_\text{neg}=\{\hat{o}_i^\text{neg}\}$  from $O_\text{neg}$  to maintain failure diversity,  and $\hat{O}_\text{pos}=\{\hat{o}_i^\text{pos}\}$  from $O_\text{pos}$  to prioritize higher-quality successful traces. The final batch used for policy updates contains $G$ rollouts, where $|\hat{O}_\text{pos}| +|\hat{O}_\text{neg}| = G$.

\begin{itemize}
\item \textbf{Negative  samples: preserving diversity}. For zero-reward rollouts ${O}_{\text{neg}}$, we apply no filtering and sample $\hat{O}_\text{neg}=\{\hat{o}_i^\text{neg}\}$  equal to half of the original group (i.e., $\lfloor\frac{|{O}_\text{neg}| }{2}\rfloor$), following their original distribution. This ensures that the model is exposed to a wide range of failure modes and learns to avoid varied error patterns.
\item\textbf{Positive samples: filtering environment noises and promoting  higher quality}. For successful rollouts $O_\text{pos}$ with a final outcome reward of 1, we sample half of the trajectories, prioritizing higher-quality traces to reinforce more effective reasoning. Specifically, each trajectories is scored for whether it contains  two types of  intermediate issues: \textit{(i)} tool call errors and \textit{(ii)} format violations.  
For tool call errors, we track the three failure modes described in Section~\ref{sec:toolerros}. For each trajectory, we  count the total number of tool calls and the number of errors, then compute a tool error ratio $p_\text{err}$.  Trajectories without tool calls are assigned a default  $p_\text{err}=0.5$ to encourage tool usage: 
\begin{align}
	\quad p_\text{err}=\begin{cases} 0.5 &\text{if no tool calls}, \\ \frac{\text{num of error tool calls}}{\text{num of all tool calls}} & \text{otherwise}. \end{cases}
\end{align}
In addition to direct coding tool errors, we observed that multi-turn rollouts in the coding environment can easily produce undesirable formats, such as redundant  $<$\texttt{reason}$>$ blocks appearing after the  $<$\texttt{answer}$>$ block. To address this, we deprioritize rollouts that violdate structural constraints. 
Specifically, we check  the number of  $<$\texttt{answer}$>$ tags. Trajectories with no tag receive the maximum downsample weight, while those with multiple tags (often causing repetition) are penalized proportionally: 
\begin{align}
	\quad p_\text{format}=\begin{cases} 1 &\text{if no $<$answer$>$ tags}, \\ min(1, \frac{\text{num of $<$answer$>$ tags}-1}{\text{num of turns}}) & \text{otherwise}. \end{cases}
\end{align}

The total penalty score of each trajectory is computed as  $p_\text{total} = p_\text{err} + p_\text{format}$. We then sample half of the positive rollouts with probability  \textit{inversely} proportional to $p_\text{total}$, so lower-penalty trajectories are more likely to be selected. This strategy guides the model toward higher-quality trajectories with correct tool usage and clean formatting, while maintaining exposure to diverse successful behaviors.
\end{itemize}

To this end, we introduce our final RL objective, GRPO-RoC, formulated as follows: 	
\begin{align}
	J_{\text{GRPO-RoC}}(\theta)
	&= \mathbb{E}_{(q,a) \sim {D},\, \{o_i\}_{i=1}^{\textcolor{red}{2G}} \sim \pi_{\theta_{\text{old}}}(\cdot|q)} \notag \\
	&	\left[
	\frac{1}{\sum_{i=1}^{G}|\textcolor{red}{\hat{o}_i}|}\sum_{i=1}^{G}\sum_{t=1}^{|\textcolor{red}{\hat{o}_i}|} \left( \min\left[\frac{\pi_\theta(\textcolor{red}{\hat{o}_{i,t}}|q,\textcolor{red}{\hat{o}_{i,<t}})}{\pi_{\theta_{old}}(\textcolor{red}{\hat{o}_{i,t}}|q,\textcolor{red}{\hat{o}_{i,<t}})} \textcolor{red}{{\hat{A}}_{i,t}},\; \mathrm{clip}(\frac{\pi_\theta(\textcolor{red}{\hat{o}_{i,t}}|q,\textcolor{red}{\hat{o}_{i,<t}})}{\pi_{\theta_{old}}(\textcolor{red}{\hat{o}_{i,t}}|q,\textcolor{red}{\hat{o}_{i,<t}})}, 1 - \textcolor{red}{\varepsilon_\text{low}}, 1 + \textcolor{red}{\varepsilon_\text{high}})\,  \textcolor{red}{{\hat{A}}_{i,t}} \right] \right)\right]\notag \\
	& \text{s.t.} \quad \{\textcolor{red}{\hat{o}_i}\}^{G}_{i=1}\in\{o_i\}^{\textcolor{red}{2G}}_{i=1} \quad \text{are sampled via RoC}.\\
	&\text{where}\quad  \textcolor{red}{{\hat{A}}_{i,t}}=\frac{\hat{r}_i-\text{mean}(\{\hat{r}_1,\hat{r}_2,\cdots,\hat{r}_G\})}{\text{std}(\{\hat{r}_1,\hat{r}_2,\cdots,\hat{r}_G\})}
	\label{eq:our_loss}
\end{align}
\textcolor{red}{$2G$} denotes the oversampled rollout trajectories, \textcolor{red}{$\hat{o}_i$} represents those selected via RoC sampling, and $\hat{r}_i$ is the 0-1 answer reward for  rollout \textcolor{red}{$\hat{o}_i$}. The clipping thresholds \textcolor{red}{$\varepsilon_\text{low}$} and \textcolor{red}{$\varepsilon_\text{high}$}   are hyperparameters, set to 0.2 and 0.28 respectively, following the Clip-Higher strategy. 

As shown in Fig.~\ref{fig:tcerror},  under GRPO-RoC, the coding tool errors within positively rewarded trajectories decreases significantly for both Qwen3-14B-base and Qwen2.5-32B-instruct. Furthermore, as shown in Fig.~\ref{fig:roccompare}, the  reduction in tool call errors  leads to significant improvements in reasoning performance and shorter, more concise responses.
These results show that GRPO-RoC simultaneously strengths reasoning capabilities and improves tool-use proficiency, resulting in smarter agentic reasoning overall. More broadly, this highlights a central value of agentic reinforcement learning by demonstrating that models can actively learn from and adapt to the external environment.

	\section{Large-Scale Agentic RL Infrastructure}
\begin{figure*}[t]
\centering
\includegraphics[width=1\textwidth]{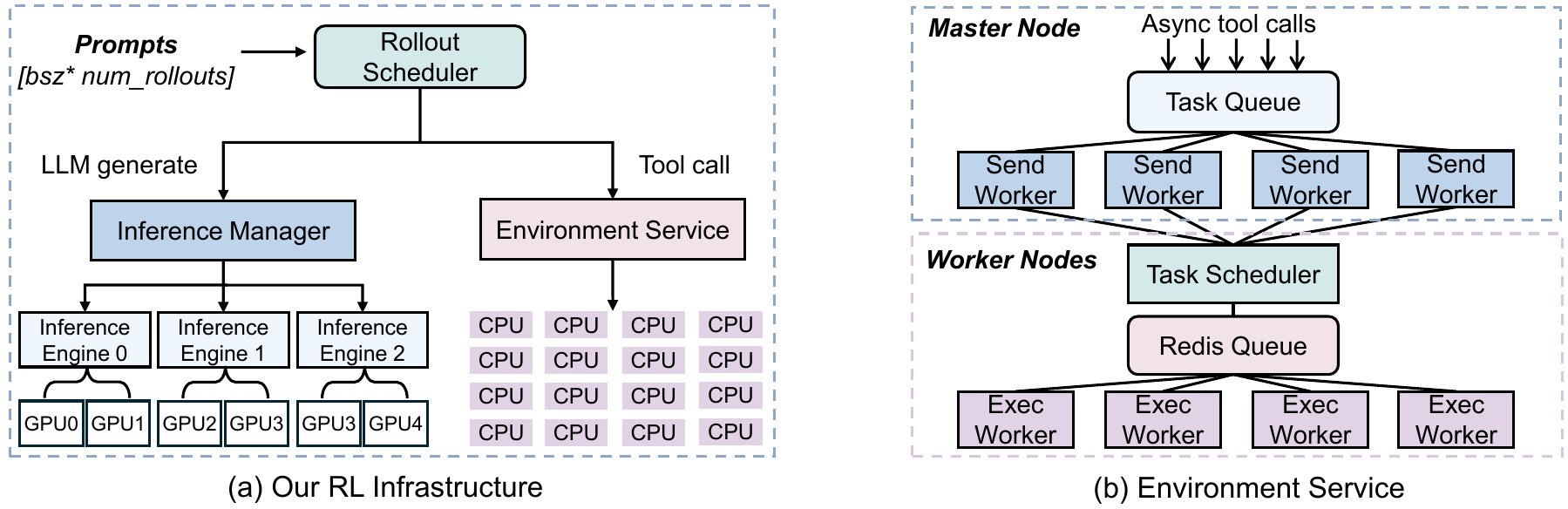}	
\vspace{-4ex}
\caption{The overall design of our agentic reinforcement learning infrastructure.}	
\label{fig:rlinfra}
\end{figure*}

	Agentic reinforcement learning introduces significant infrastructure challenges. To enable  large-scale training, we build a custom agentic RL infrastructure on top of VERL v0.2~\citep{verl} and SGLang~\citep{sglang}, as shown in FIg.~\ref{fig:rlinfra}. Specifically, we address two major bottlenecks:
	
	\begin{itemize}
\item \textbf{Massive Concurrent Tool Calls}.  A naive approach to obtaining coding tool outputs is to execute the generated code directly using a local Python interpreter. However, in large-scale multi-turn rollouts, a single training batch can trigger thousands of code execution requests. Running all these tool calls locally not only overwhelms CPU resources but also leaves GPUs idle, significantly slowing rollout speed as shown in Fig.~\ref{fig:rollout}. More critically, LLM-generated code is unpredictable and may contain bugs, uncontrolled threads, or hard-to-kill external library calls, posing a severe risk to the main training process.  To address both efficiency and safety, we implement a dedicated, isolated code environment service capable of handling massive concurrent tool call requests without stalling rollouts.
\item \textbf{Highly Imbalanced Multi-turn Rollouts}. In standard RL training, rollouts in a  batch are statically and evenly assigned to GPUs, but differing response lengths leave many GPUs idle while waiting for the longest rollout, leading to poor GPU utilization and slow training. This problem is amplified in agentic RL, where each response spans multiple turns of uneven token generation and tool calls. When scheduled statically and synchronously, these imbalances recur at every turn, compounding worst-case latency and increasing idle time. To address this, we introduce a load-balanced rollout scheduler that dynamically allocates rollout requests based on available KV cache capacity  across GPUs. 
\end{itemize}

\subsection{Reliable High-Throughput Code Environment}

\begin{figure*}[h]
\centering
\includegraphics[width=0.7\textwidth]{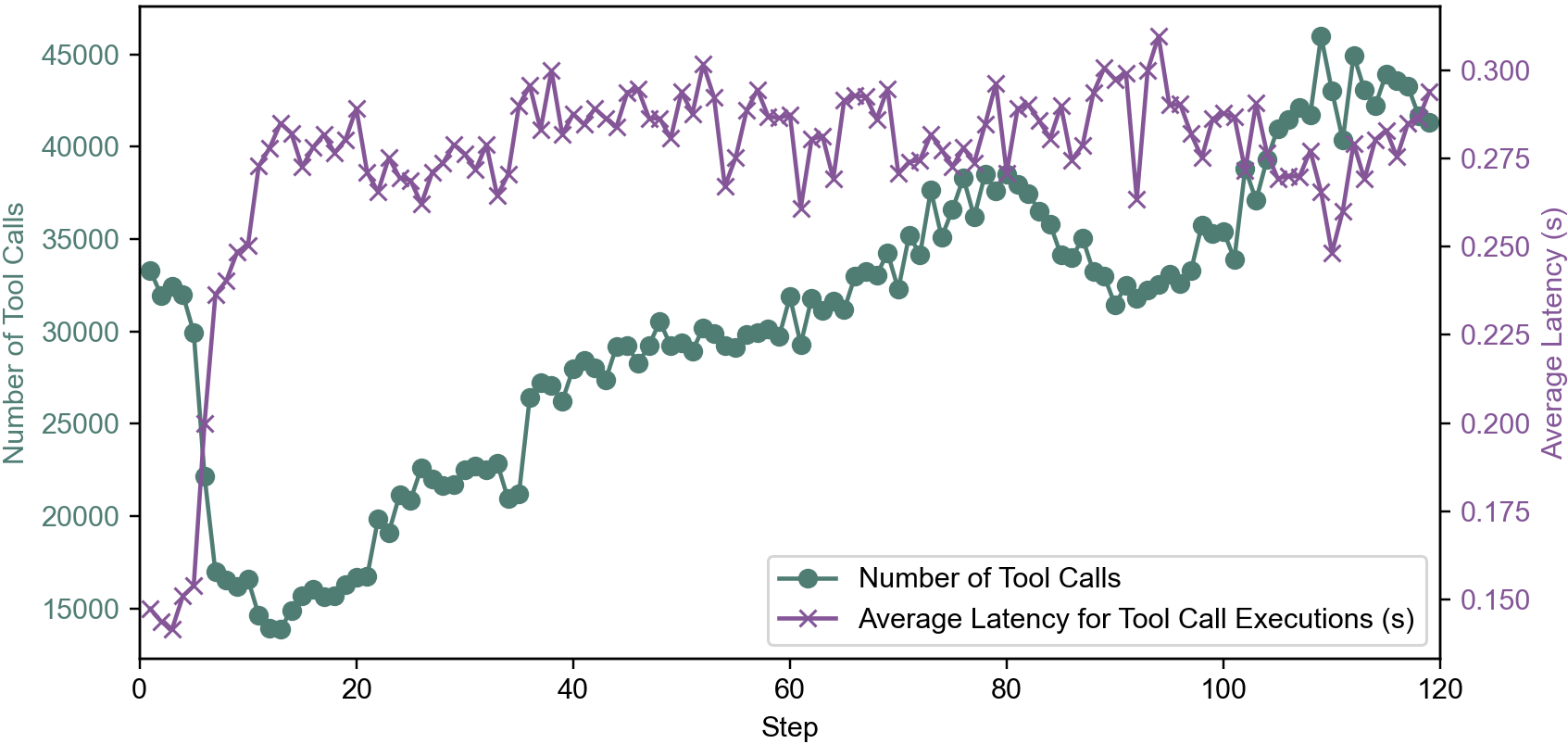}	
\vspace{-2ex}
\caption{Our code environment demonstrates scalability by reliably handling up tp 45K concurrent tool calls per step, while maintaining consistently low end-to-end latency from dispatch to response.   }
\label{fig:toolcalls}
\end{figure*}

Fig.~\ref{fig:rlinfra}(b)  shows the design of our environment service, which is developed with two main objectives. The first is to isolate the service from the main RL training process while maximizing resource utilization. The second is to support a large number of concurrent tool calls and return execution results as quickly as possible.

The service is distributed across CPU cores of our 64 AMD MI300X GPU training cluster.  On the master node, a centralized \texttt{task queue} along with 32 \texttt{send workers} manages the dispatch of tool call executions. The remaining worker nodes each run a lightweight \texttt{task scheduler} and a pool of 1024 \texttt{execution workers} to perform the actual tool call execution. To handle  massive concurrent tool calls, each request is added to the centralized \texttt{task queue} to avoid overloading the workers. The 32 \texttt{send workers} continuously poll this queue, grouping up to 64 tool calls into a batch. A batch is dispatched either when it reaches capacity or after a fixed timeout, and the \texttt{send worker} waits for execution results before sending the next batch. On the worker nodes, the \texttt{task scheduler} dynamically assigns tool calls from incoming batches to idle \texttt{execution workers}, ensuring balanced workload distribution. Once execution is complete, results are returned to the \texttt{send workers}, which forward them back to the RL rollout process. This architecture ensures isolated, efficient, and reliable code environment at large scale.

To evaluate the effectiveness of our environment service, we measure the average latency from issuing a tool call to receiving its result. As shown in Fig.~\ref{fig:toolcalls}, each training step can generate up to 45K tool calls. Even at this scale, the service achieves both high throughput (45 calls per step) and low latency (0.3 seconds per call, including scheduling and execution time), demonstrating its ability to support large-scale training without becoming a bottleneck.

\noindent\textbf{Extended Functionality: Answer Correctness Verification}. In our experiments, we find that rule-based reward systems such as the \texttt{Math-Verifier} can occasionally take a long time to run, especially on complex or edge-case extracted math answers. Running these verifications directly in the training loop can block rollouts progress and causes  GPU idle time. To avoid this, we offload answer verification to the environment service, allowing these CPU-intensive computations to run asynchronously without stalling training.

\subsection{Load-Balanced Rollout Scheduler}

\begin{figure*}[t]
\centering
\includegraphics[width=1\textwidth]{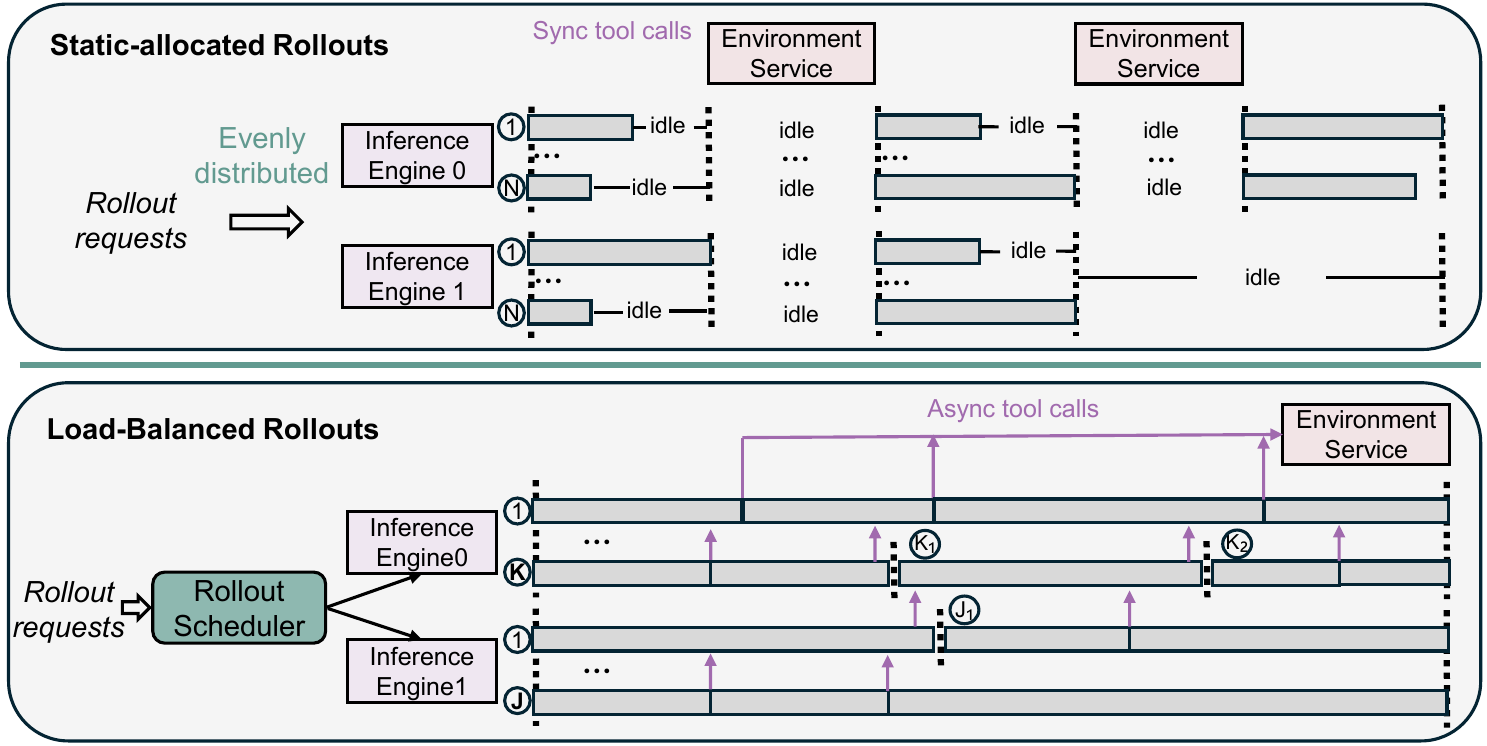}	
\vspace{-2ex}

\caption{\textit{Top}: Naively static rollout allocation leads to significant GPU idle time and synchronization delays.   \textit{Bottom}: our dynamic load-balanced scheduler that assigns rollouts based on available KV cache, dispatches tool call execution asynchronously, and balances computation across GPUs. For example, $K_1$, $K_2$, $J_1$ denote the number of rollouts computed from the current available KV cache memory on inference engines 0 and 1. }	
\label{fig:rollout}
\end{figure*}

\noindent\textbf{Static Rollout Allocation:  Load Imbalance,  Synchronization Delays and KV Cache Overflow}. Rollout inefficiency is a well-known challenge in RL training infrastructure, and it becomes even more pronounced in agentic RL,  where each responses consists of multiple turns of token generation and numerous tool calls, creating high variability in computational load. In our early implementation, we built the rollout system on top of VERL v0.2 using a  straightforward \textit{statically allocated batch inference} strategy. As shown in Fig.~\ref{fig:rollout} (upper), VERL evenly  pre-allocates all rollout requests across GPUs, with each GPU receiving $N$ rollouts. However, this static allocation fails to manage the significant variability in computation across multi-turn rollouts, leading to several key inefficiencies.

First, despite GPUs being statically assigned the same number of rollout requests, the total computational workload across GPUs can be highly imbalanced. Each rollout may have a different number of turns, and each turn can vary in token length. These turn-level token length imbalance repeatly create GPU idle time, as shorter rollouts must wait for the longest rollout within each turn to complete.  Moreover, synchronization delays from tool calls, which are typically collected and executed together per turn, further increase idle time, Together, these factors under static allocation lead to severe GPU-level workload imbalance and substantial idle time.

Second, static rollout allocation can trigger KV cache overflow, which further reduces rollout efficiency. Inference engines like SGLang cannot predict in advance how many tokens each rollout will generate,  so  all assigned rollout requests are launched in parallel by default.  When a GPU's KV cache exceed its capacity, SGLang evicts half of the in-progress rollouts, even if partial computation has already been completed. The evicted rollouts must then be recomputed after the remaining rollouts finish, resulting in significant wasted computation.

\noindent\textbf{Dynamic Load-Balanced Rollout Scheduling}. To address these challenges, we introduce a \textit{load-balanced rollout scheduling} method, as  illustrated in Fig.~\ref{fig:rollout} (bottom). The design principle is to dynamically allocate rollout requests to maintain balanced total computation across GPUs,   while avoiding any wasted computation from KV cache overflow and recomputation. 
As shown in Fig.~\ref{fig:rollout} (bottom), our dynamic rollout scheduler assigns requests based on the current available KV cache capacity of each GPU rather than statically dividing them evenly. Specifically, given a maximum rollout length $L$, we estimate the maximum number of rollouts $K$ ($K<N$) that each GPU can safely handle without exceeding its KV cache limits. Each GPU then executes its assigned rollouts independently. During multi-turn rollouts, tool calls are dispatched asynchronously to the environment service immediately upon generation, eliminating idle time caused by waiting for other rollouts.  Once a GPU finishes the assigned requests and frees KV cache space, the scheduler assigns new requests in real time, ensuring balanced workloads across GPUs. This approach  significantly improves GPU utilization and overall rollout efficiency.

\section{Training Recipe}
This section presents our recipe for advancing {\sysname}-14B at minimal compute scale, covering dataset curation, multi-stage training and lessons from unsuccessful attempts.

We use Qwen3-14B-base~\citep{qwen3technicalreport} as our base model. To achieve frontier-level performance with minimal compute, training begins with a  \textit{non-reasoning SFT} stage followed by multi-stage efficient RL with progressively increasing  training lengths.  Specifically, the non-reasoning SFT enables the model initially produces relatively short responses, while multi-stage RL with  GRPO-RoC further shortens response length throughout RL and significantly reduces computational requirements.

\subsection{Non-Reasoning Cold Start for Instruction Following}
\label{sec:sft}
\begin{table}[ht]
	\centering
	\caption{Performance of Qwen3-14B-base after our non-reasoning SFT. The model improves on tool use, instruction following, and chat, while  maintaining comparable math reasoning ability to the base model.}
	\vspace{-1ex}
	\label{tbl:sft}
	\resizebox{0.98\textwidth}{!}{
		\begin{tabular}{ccccccc}
			\toprule
			\multirow{2}{*}{Model} & \multicolumn{3}{c}{Math Reasoning} & Tool & Instruction following & Chat\\
			\cmidrule{2-4} 
			& MATH-500 &AIME24&AIME25&BFCL v3& IFEval$_\text{strict prompt}$ &Arena-Hard\\ \midrule
			Qwen3-14B-Base& 62.0& - & - & - & - &-\\
			Qwen3-14B&96.8 & 79.3& 70.4& 61.5 & 84.8&86.3\\
			\midrule 
			Our non-reasoning SFT & 57.4& 3.33&0 &63.1 &83.7 &86.8\\
			\hline
		\end{tabular}%
	}
\end{table}

Unlike prior work~\citep{liu2025prorl,retool,qwen3technicalreport} that includes heavy reasoning data in SFT, we focus solely on   general instruction-following, JSON formatting, and basic coding tool usage, which are essential for agentic RL.  
We incorporate the following datasets: \textbf{(1)} 165K function call data, including 117K from ToolACE-11K~\citep{liu2024toolacewinningpointsllm}, APIGen-MT-5K~\citep{prabhakar2025apigen}, Glaive-function-calling-v2-101k~\citep{glaive}, along with  48k Magicoder datasets~\citep{wei2023magicoder} reformatted into  JSON function call format to enhance   coding tool capabilities. \textbf{(2)} 30K instruction-following examples from Tulu3 post-training dataset~\citep{ifdata}, with response rewritten using   o4-mini to improve quality. \textbf{(3)} 27K chat data   from LLaMA-Nemontron post training dataset~\citep{bercovich2025llama}, with prompts for each conversation rewritten using o4-mini. Table~\ref{tbl:sft} shows the performance across different capabilities after SFT. As reported, our non-reasoning SFT primarily improves the  base model's tool use,  instruction-following and chat abilities, while maintaining comparable math  performance to the base model.

\subsection{RL Data Curation}
To ensure reliable RL supervision, we follow two rules when collecting math problems. First,
problems must be high-quality, challenging, and have correctly labeled final answers. Second,  answers must be integers. The integer-only requirement is is essential because verifying equivalence between different algebraic expressions is notoriously difficult. For example, current rule-based verifiers such as \texttt{Prime} and \texttt{math\_verifier} struggle to recognize that \texttt{(a+b)(b+c)(c+a)} and \texttt{(a+c)(c+b)(b+a)} represent the same solution. Such cases can lead to incorrect rewards by misclassifying correct rollouts as incorrect.  
Therefore, we only include math problems with integer answers.

Guided by these rules, we collect over 100K candidate problems from three sources. First, we include 17K integer-only problems from the DAPO training set~\citep{dapo}.  Next, we add 93K problems from the Art of Problem Solving (AoPS) forums  via  OpenMathReasoning~\citep{openmathreasoning}. Finally, we include 937 challenging problems from Project Euler~\citep{eluer}, which require both mathematical insight and programming skills.

We then perform extensive cleaning to produce a final set of \textbf{42K} high-quality problem-answer pairs. For the OpenMathReasoning dataset, we remove unverifiable answers (e.g., \texttt{The limit does not exist}), overly complex formats (e.g., \texttt{Perimeter=54cm, Area=180cm²}), and incorrect answers. Specifically, we use Qwen3-32B to generate 16 responses per problem and retain only those with integer answers that match the original labeled answer at least twice. For the Project Euler dataset, we remove problems with excessively large numerical answers (e.g., \texttt{6.5e27330467}) that can cause verifiers to time out. This process produces a clean, verifiable dataset for RL training.

\subsection{Multi-Stage RL Training} 
\label{sec:rltraining}

We then run large-scale agentic reinforcement learning using GRPO-RoC, with a   learning rate of 1e-6 and a batch size of 512 prompts. 
For each prompt, we first oversample $2G=32$ multi-turn rollouts and then select 16 using the RoC strategy. To improve training efficiency, we adopt a multi-stage strategy that gradually increases both the maximum training length and the difficulty of the data as shown in Table~\ref{tbl:rltrainingrecipe}. Unlike other RL methods that heavily rely on scaling the training length, often using at least 16K token length throughout, we start with shorter lengths and scale up across stages (8K$\rightarrow$12K$\rightarrow$12K). This high efficiency is enabled by GRPO-RoC, which allows strong performance even with shorter response length. We detail each stage below.

\begin{table}[th]
	\centering
	\caption{Comparison of training recipes among leading reasoning models. {\sysname} is trained with non-reasoning  SFT, uses much \textbf{short  RL training lengths}, and applies data difficulty filtering only at the final stage.}
	\vspace{-1ex}
	\label{tbl:rltrainingrecipe}
	\resizebox{0.98\textwidth}{!}{
			\begin{tabular}{@{\hskip0pt}c@{\hskip4pt}c@{\hskip4pt}c@{\hskip4pt}c@{\hskip4pt}c@{\hskip4pt}c@{\hskip0pt}}
				\toprule
				Method&Has Reasoning SFT?&RL Stages&Total Steps& Max Training Length & Data Difficulty Filtering\\
				\hline
				DeepSeek-R1-Zero&\xmark& -& $>$9K &-&-\\
				DeepSeek-R1&\cmark& -& - &-&-\\
				DAPO~\citep{dapo} &\xmark&1 & $>$5K&20K &\xmark\\
				ReTool~\citep{retool} & \cmark & 1 & 400& 16K& -\\
				MiniMax~\citep{chen2025minimax} & \cmark & 4&$>$4K & 40K$\rightarrow$48K$\rightarrow$56K$\rightarrow$80K& All Stages\\
				MiMo~\citep{xiaomi2025mimo} & \cmark&3 & 175K&32K$\rightarrow$38K$\rightarrow$48K& All Stages\\
				Magistral~\citep{rastogi2025magistral}&\cmark& 3&- &16K$\rightarrow$24K$\rightarrow$32K&All Stages\\
				\midrule 
				\bf {\sysname} & \xmark &\bf  3& \bf 510 & \bf 8K$\rightarrow$12K$\rightarrow$12K& \bf Only Stage 3\\
				\bottomrule
			\end{tabular}%
		}
	\end{table}

	\noindent\textbf{RL Stage-1: Concise Training at 8K Response Length}. 
	In the first stage, we train on the full set of 42K curated math problems using a maximum response length of 8K tokens. This shorter maximum length is feasible because, after the non-reasoning SFT, the model initially produces relatively short responses (around 1K tokens, as shown in Fig.~\ref{fig:multistage}). Combined with GRPO-RoC, which provides more efficient and effective reasoning capabilities, this ensures that the model response length remains moderate throughout early RL training.

	As shown in Fig.~\ref{fig:multistage}(c), during Stage 1 the average response length starts at around 1K tokens and gradually increases, eventually stabilizing at approximately 4K tokens. During this period, the clipping ratio, defined as the fraction of rollouts exceeding the 8K limit, temporarily surpasses 10\% multiple times. While such high clipping ratio often suggest that the maximum response length may be insufficient, we keep the 8K limit. This encourages the model to better utilize GRPO-RoC to reason more effectively, and we observe that it quickly self-adjusts. The clipping ratio decreases over subsequent steps, evaluation scores improve, and responses become concise. These findings show that concise training under a shorter length budget not only improves training efficiency but also promotes stronger reasoning early on, laying a solid foundation for later stages of multi-stage RL.

	\noindent\textbf{RL Stage-2: Extending to 12K Response Length}. By the end of Stage 1 (i.e., 300 steps), the rollout clipping ratio stabilizes around 10\%, and both training rewards and evaluation scores plateau. This suggests that although the model reasons more effectively, 
	the 8K maximum response length has become a limiting factor for further learning.  
	In Stage 2, we therefore increase the maximum response length to 12K tokens. As shown in Fig.~\ref{fig:multistage}, this extension increases the average response length from 4K to 6K and yields consistent improvements on AIME24 and AIME25.

	\noindent\textbf{RL Stage-3: Focused Training on Difficult Problems}. By the end of Stage 2, over 70\% of problems in a batch are rejected due to  achieving a perfect pass rate of 1, showing that many problems have now become too easy for the model. To maintain training effectiveness, we shift focus to harder problems in Stage 3. Unlike prior approaches that dynamically exclude perfectly solved problems during training, we adopt an offline filtering strategy. Specifically, we use the latest policy (from the final 385 steps of Stage 2) to generate 8 rollouts per problem on the original 42K set and remove problems with all 8 correct. This filtering yields 17.3K harder problems. For training on this dataset, we reset optimizer states and update the reference model the latest policy.

	As shown in Fig.~\ref{fig:multistage}, focusing on harder problems further improves performance and  increases the average response length from 6K to 8K. Over an additional 125 steps, this stage gradually advances the 14B model to frontier-level mathematical reasoning. Beyond this point, performance begins to saturate and can even decline, so we stop at 125 steps (see the discussions in Sec.~\ref{sec:ablate}).  In total, the three RL stages comprise 510 training steps, all completed on 64 MI300X GPUs within one week.

	\subsubsection{Unsuccessful Attempts And Lessons}
	In the early development of {\sysname}-14B, we experimented with various RL training approaches and faced several challenges. We share these experiences to provide practical insights, rather than to imply that these methods are inherently ineffective for building strong reasoning models.

	\noindent\textbf{Overlong filtering further increases rollout truncation}. In RL training, rollouts exceeding the maximum response length are truncated and assigned a negative reward, since they fail to reach a final answer. 
	DAPO~\citep{dapo} suggests that penalizing such response can confuse the model, since well-reasoned rollouts may be unfairly penalized simply for its excessive length.  To address this, DAPO proposes overlong filtering, which discards truncated rollouts entirely without assigning reward, a strategy later adopted in several follow-up works~\citep{liu2025acereason,deepcoder2025}. In our experiments, however, overlong filtering yields no  benefits. On the contrary, it increases the ratios of overlong rollouts.  One possible reason is that many of these overlong responses contain repetitive patterns. Without negative feedback, the model receives no signal to correct them and continues producing excessively long outputs.

	We therefore do not adopt overlong filtering. Instead, we keep truncated rollouts with negative reward,  which turn out to be useful training signals. These rollouts guide the model to reduce repetition and adapt its behavior. When the clipping ratio spikes, the model quickly adjusts in subsequent steps, bringing the ratio back to a reasonable level.

	\noindent\textbf{N-gram repetition detection risks removing effective reasoning patterns}. Beyond our current rollout sampling strategy, we also explored finer-grained scoring of intermediate behaviors in trajectories. Specifically, we  experimented with lowering the sampling probability of correct rollouts that exhibit repetition patterns,  as part of our resample-on-correct strategy. We follow the n-gram repetition detection method used in Phi-4-Reasoning~\citep{abdin2025phi}. However, in our experiments, this approach negatively affected both the model's average response length and its reasoning scores. With closer analysis of the filtered repetitive rollouts, we find that it is inherently difficult to precisely distinguish between undesirable repetition and legitimate reasoning behavior. For example, the model may generate two similar tool calls with different inputs to verify its results. While this behavior reflects thoughtful reasoning and is desirable,  it is often incorrectly flagged as repetition by simple n-gram heuristics.

	From these experiences, we draw a broader lesson about reward design. LLM RL is inherently self-exploratory,  with highly diverse and unpredictable intermediate behaviors.  Overly complex, rule-based rewards or scoring schemes can introduce bias, penalize useful behaviors, and fail to generalize across reasoning patterns. To address this, we adopt a \textbf{minimal reward design} based solely on final answer correctness. Other low-quality intermediate behaviors, such as environment noises or formatting issues, are addressed via resample-on-correct rollout strategy rather than direct penalties. This approach reduces bias, preserves exploration, and ensures robust learning throughout training.

\section{Experiments}
	\subsection{Setup}
\noindent\textbf{Training Setup}. We run experiments on two models: Qwen3-14B-Base and Qwen2.5-32B-Instruct. For Qwen3-14B-Base, we preform a non-reasoning SFT before RL to instill basic tool-use and instruction-following abilities, as described in Sec.~\ref{sec:sft}. The SFT is trained for 3 epochs with a learning rate of 5e-6, 4\% warm-up steps, cosine  decay, and a batch size of 128. For Qwen2.5-32B-Instruct, no additional SFT is applied. For RL training, we use the AdamW optimizer with a constant learning rate of 1e-6 and linear warm-up over 20 rollout steps. We use a rollout temperature of 1.0 and set the maximum number of multi-turn rollouts  to $T$=10 for the first two RL stages and $T$=15 in the final stage.
All experiments are conducted on 64  AMD MI300X GPUs. For Qwen2.5-32B-Instruct, we include experiments to enable fair comparison with prior representative RL works (e.g., DAPO, ReTool), which are also conducted at Qwen2.5-32B scale. Due to limited resources, we only run stages 1 and 2 for this model.

\noindent\textbf{Evaluation Benchmarks}. Although our {\sysname}-14B is RL-trained solely on math data, we evaluate it across diverse domains to assess the general effectiveness of our approach: 
 \textbf{(i)} Competitive math benchmarks, including \textbf{MATH-500}~\citep{lightman2023let}, \textbf{AIME24} and \textbf{AIME25}~\citep{aime}, and \textbf{HMMT25}~\citep{matharena}. To ensure fair and unbiased evaluation, we decontaminate our training data by removing any problems with 8-gram overlaps  against these benchmarks; \textbf{(ii) GPQA-Diamond}~\citep{rein2024gpqa}, for evaluating general reasoning and scientific problem-solving; \textbf{(iii) BFCL v3}~\citep{bfcl}, for evaluating agentic tool use capabilities; and \textbf{(iv) IFEval}~\citep{zhou2023instruction} and \textbf{Arena-Hard}~\citep{li2024crowdsourced}, for measuring general alignment performance. 

We use task-specific inference settings. 
For math benchmarks and GPQA-Diamond, we allow up to 30K tokens per response with a temperature of 0.6, applying the prompt template in Fig.~\ref{fig:prompt} with a maximum of  $T=30$ turns. Each question is sampled 16 times, and we report  average pass@1 accuracy and response length in tokens. For BFCL v3, IFEval, and Arena-Hard, we use each benchmark's default prompt template with a temperature of 0.

\subsection{{\sysname}-14B Main Results}

\begin{table}[t]
	
	\caption{With GRPO-RoC agentic RL training, {\sysname}-14B achieves competitive mathematical reasoning comparable with frontier LLMs, while using significantly less training compute and smaller model sizes. }
	\vspace{-1ex}
	\label{tbl:mainresults}
	\resizebox{0.98\textwidth}{!}{
			\begin{tabular}{lcccccc}
				\toprule
				Model             &Reasoning SFT before RL?         &MATH-500           & AIME24               & AIME25 & HMMT Feb.25\\ \midrule
				OpenAI	o3-mini (medium)&- & 98.0& 79.6& \bf 77.0& \bf 53.0 & \\
				DeepSeek-R1 (671B)&\cmark& 97.3& 79.8& 70.0&44.4\\
				Claude-Opus-4.0 (Think) &\cmark& \bf 98.2& 76.0 & 69.2&-\\
				Kimi k1.5& \cmark&96.2& 77.5&-&-\\
				Magistral Medium &\cmark&94.3 & 73.6& 64.9& -\\
				QWQ-32B & \cmark& 98.0&79.5 & 65.8&47.5\\

				Magistral Small (24B) & \cmark & 95.9 & 70.7& 62.8&35.7\\
				Qwen3-14B &\cmark&96.8 &79.3& 70.4 &48.9 \\
				
				DeepSeek-R1-Zero (671B) &\xmark& 95.9 & 71.0 & 53.3 &46.0 \\
				\rowcolor{lightgreen}	\bf{\sysname}-14B &\bf\xmark&\bf 97.8 &\bf 80.6 &\bf 69.8 &\bf 52.7\\
				\hline
			\end{tabular}%
		}
	\end{table}

		\begin{figure*}[t]
		\centering
		\includegraphics[width=1\textwidth]{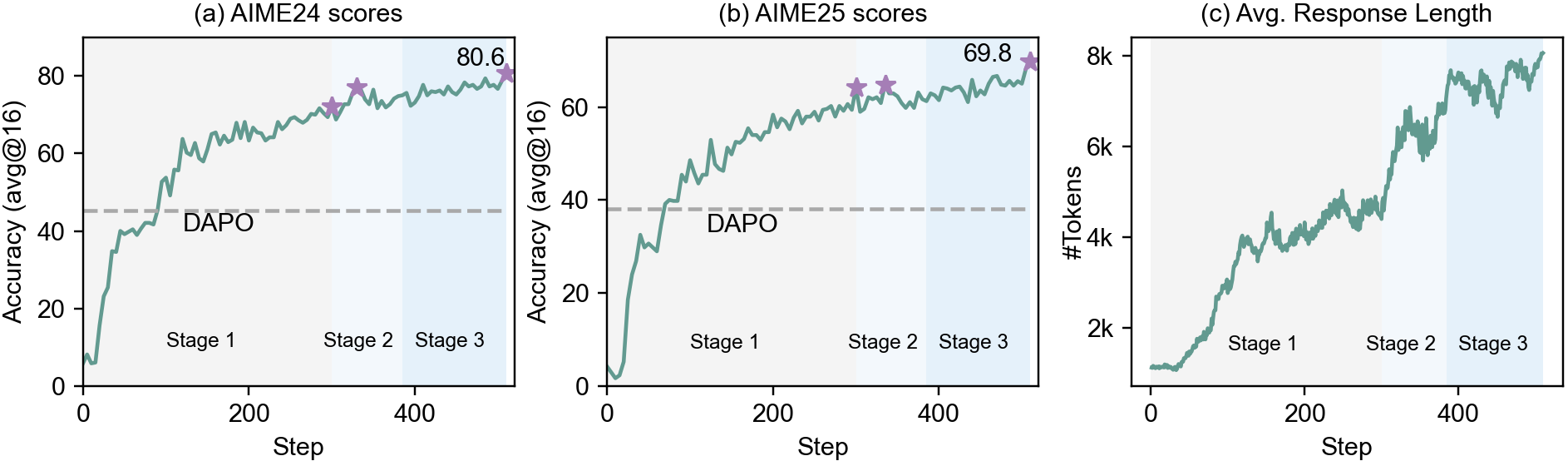}	
		\vspace{-3ex}
		\caption{AIME24/AIME25 accuracy and average training response lengths throughout multi-stage RL training.  }	
		\label{fig:multistage}
	\end{figure*}

	\noindent\textbf{Competitive math reasoning from pure agentic RL with minimal compute}. Table~\ref{tbl:mainresults} summarizes the final mathematical reasoning performance of {\sysname}-14B compared to state-of-the-art reasoning models. We highlight two key observations: \textbf{(i)} {\sysname} substantially boosts a 14B pre-trained model to state-of-the-art levels, matching and even surpassing more heavily and much larger trained frontier LLMs. On AIME24, {\sysname}-14B achieves an average accuracy of 80.6\%, outperforming o3-mini (medium), DeepSeek-R1, and Claude Opus 4.0 (thinking) by 1.0\%, 0.8\%, and 3.6\%, respectively. On AIME25 and HMMT25, it reaches 69.8\% and 52.7\%, demonstrating consistently strong performance across benchmarks.
	\textbf{(ii)} Effective agentic RL alone yields surprisingly strong reasoning, outperforming state-of-the-art zero-RL baselines. As shown in Table~\ref{tbl:mainresults}, most frontier models rely on reasoning-specific SFT to warm-start the policy, whereas {\sysname} uses only a lightweight, non-reasoning SFT for tool formatting and instruction following. Despite this minimal setup, GRPO-RoC boosts performance \textit{from near-zero to 80.6\% on AIME24 and 69.8\% on AIME25} (pass@1). Moreover, compared with zero-RL models such as DeepSeek-R1-Zero, {\sysname}-14B delivers substantially stronger results across all benchmarks, demonstrating the power of agentic RL as a standalone driver of advanced reasoning.  
	These results
	  are especially notable given the small 14B scale and highly cost-effective training compute (e.g., 510 RL steps on 64 MI300X GPUs).  Unlike large-scale efforts that rely on extensive data and compute budgets, {\sysname} delivers state-of-the-art reasoning with comparatively lightweight training, highlighting a practical path toward efficient reasoning model development.

	\noindent\textbf{Per-RL stage improvement}. To understand how {\sysname}-14B achieves its strong performance, we show the step-by-step improvements and average training lengths across the three RL training stages. 
As shown in	Fig.~\ref{fig:multistage}(a,b),  math reasoning performance on AIME24 and AIME25 steadily improves across stages. Stage 1, with concise RL training and an 8k max response length, already yields substantial gains. AIME24 improves from 3.3\% (SFT) to 72.1\% and AIME25 from 0\% to 64.2\%, surpassing the CoT-only DAPO baseline by +21.7\% and +21.3\% respectively. Stage 2, enabled by a 12k max response length, further increases scores to 77.0\% (AIME24) and 64.8\% (AIME25). Stage 3, training on harder problems, boosts performance to 80.6\% and 69.8\%. 
	
\noindent\textbf{Smarter reasoning with fewer tokens.} {\sysname} not only achieves strong reasoning but also enables more effective reasoning with fewer  tokens.  Table~\ref{tbl:responselength} shows the average response length on the AIME24 and AIME25 benchmarks, comparing {\sysname}-14B with DeepSeek-R1-Zero, QWQ-32B and the official Qwen3-14B. Despite generating shorter responses, {\sysname}-14B attains higher reasoning accuracy on these challenging problems. This indicates that, by reinforcing higher-quality positive trajectories, our model has effectively learned to use coding tools more intelligently to reason more efficiently.

	\begin{table}[t]
	\centering
	\caption{{\sysname}-14B achieves effective reasoning with significantly fewer tokens. }
	\vspace{-1ex}
	\label{tbl:responselength}
			\begin{tabular}{ccc}
				\toprule

			\multirow{2}{*}{Model}	& \multicolumn{2}{c}{The Avg. Response Length in Tokens}\\
		     & AIME24&AIME25\\
      	\midrule
			DeepSeek-R1-Zero (671B)& 14246.8 &17132.9 \\
			QWQ-32B &11868.4 &15865.4 \\
			Qwen3-14B& 14747.6& 17521.9\\
		\rowcolor{lightgreen}	\bf{\sysname}-14B & \bf 9339.7& \bf 10943.4\\
				\hline
			\end{tabular}
	\end{table}

	\begin{table}[t]
		\centering
		\caption{Despite being trained with math-only RL, {\sysname}-14B demonstrates strong performance on general tasks.  Note: scores after non-reasoning SFT are marked in gray.}
		\vspace{-1ex}
		\label{tbl:generalization}
				\begin{tabular}{ccccc}
					\toprule
				\multirow{2}{*}{Model}	     & GPQA-Diamond&BFCL  v3&IFEval$_\text{strict prompt}$ & Arena-Hard \\ 
				     & (Science Reasoning) & (Agentic Tool Use) & \multicolumn{2}{c}{(General Alignment)}\\\midrule
					DeepSeek-V3& 59.1& 57.6&\bf 86.1&85.5\\
				\rowcolor{lightgreen}	\bf {\sysname}-14B &\textbf{60.9} (\textit{\textcolor{gray}{42.1}}) & \textbf{60.8} (\textit{\textcolor{gray}{63.1}})&83.4 (\textit{\textcolor{gray}{83.7}})&\textbf{86.6} (\textit{\textcolor{gray}{86.8}})\\
					\hline
				\end{tabular}
		\end{table}

		\noindent\textbf{Strong generalization performance.} Beyond mathematical reasoning, we evaluate {\sysname}-14B on diverse benchmarks to test its generalization capabilities. As shown in Table~\ref{tbl:generalization}, after math-only agentic RL training, our {\sysname}-14B demonstrates strong generalization performance, outperforming DeepSeek-V3 on most tasks. Notably, on the science reasoning benchmark GPQA-Diamond, despite no training on science data,  {\sysname}-14B improves accuracy from 42.1\% to 60.9\%, surpassing DeepSeek-V3 by 1.8\%, showing that reasoning patterns learned from mathematics transfer effectively to general science reasoning. On non-reasoning tool-use and alignment tasks, the model shows no improvement but maintains performance comparable to our non-reasoning SFT baseline. Overall, math-only agentic RL can improve reasoning in other domains without affecting unrelated  general tasks.

	\subsection{Ablation Study and Discussions}
	\label{sec:ablate}
	
		\begin{table}[t]
		\centering
		\caption{Comparison of RL baselines with and without tools, showing {\sysname}'s consistent superiority at different model scales. On Qwen2.5-32B-Instruct, only RL stages 1 and 2 are performed due to resource constraints. }
		\vspace{-1ex}
		\label{tbl:rlbaselines}
		\resizebox{0.98\textwidth}{!}{
				\begin{tabular}{@{\hskip0pt}l@{\hskip4pt}c@{\hskip4pt}c@{\hskip6pt}c@{\hskip6pt}c@{\hskip6pt}c@{\hskip6pt}c@{\hskip0pt}}
					\toprule
					Model      &\makecell{Reasoning SFT\\ before RL?}  &Tools               &MATH-500           & AIME24               & AIME25 &RL  Steps \\ \midrule
					\textit{Qwen2.5-32B}&&&&&&\\
					DeepSeek-R1-Zero-Qwen-32B &\xmark&\xmark& 91.6  & 47.0&-&-\\
					Open-Reasoner-Zero-32B &\xmark&\xmark& 92.2 & 48.1 & 36.0&$>$1000  \\
					DAPO-Qwen-32B &\xmark&\xmark&90.3  & 50.0&32.1&$>$5000 \\ 
					VAPO-Qwen-32B&\xmark&\xmark& - & 60.4&-&$>$5000  \\
					ZTRL-32B~\citep{agentrl} &\xmark&\cmark& 87.8& 56.7&33.3  &600 \\
					ReTool-32B~\citep{retool} &\cmark&\cmark&93.4 &67.0 & 49.3 &400 \\
					\bf{\sysname}-Qwen2.5-32B &\xmark&\cmark&\bf94.8 &\bf69.4 &\bf57.3& 700\\
					\midrule 
					\textit{Qwen3-14B-Base} && & & & &\\
					DAPO-Qwen-14B~\citep{wang2025beyond}&\xmark&\xmark& 92.2&45.2&38.1&2000 \\
					DAPO-Qwen-14B w/Forking tokens~\citep{wang2025beyond} &\xmark&\xmark& 93.6&50.4 &42.9& 2000\\
					\rowcolor{lightgreen}	\bf{\sysname}-14B &\xmark&\cmark&\bf 97.8 &\bf 80.6 &\bf 69.8 &\bf 510\\
					\hline
				\end{tabular}%
			}
		\end{table}

	\noindent\textbf{Comparison with other RL approaches on the same base model}. In addition to Qwen3-14B-base, we compare {\sysname} with recent public RL methods on Qwen2.5-32B, a scale commonly used in prior works. Due to compute limits, only the first two RL stages are run, totaling 700 steps. Table~\ref{tbl:rlbaselines} presents the results. We highlight two main observations: \textbf{(i)} on both base model scales, agentic RL with coding tools consistently outperforms pure CoT-based methods, often with fewer training steps. On Qwen2.5-32B, {\sysname}, ZTRL~\citep{agentrl}, and ReTool~\citep{retool} all surpass DAPO and VAPO. Similarly, on Qwen3-14B-base, {\sysname} significantly outperforms CoT-only baselines with fewer training steps, demonstrating the effectiveness of agentic RL.  \textbf{(ii)} Compared to other agentic RL methods, {\sysname} shows clear superiority. On Qwen2.5-32B, ReTool uses reasoning-specific SFT before RL, whereas {\sysname} relies only on non-reasoning SFT. Despite this, {\sysname} achieves 2.4\% and 8.0\% higher accuracy on AIME24 and AIME25, respectively. Performance is expected to improve further with continued training in RL stage 3.

\begin{figure*}[ht]
	\centering
	\includegraphics[width=1\textwidth]{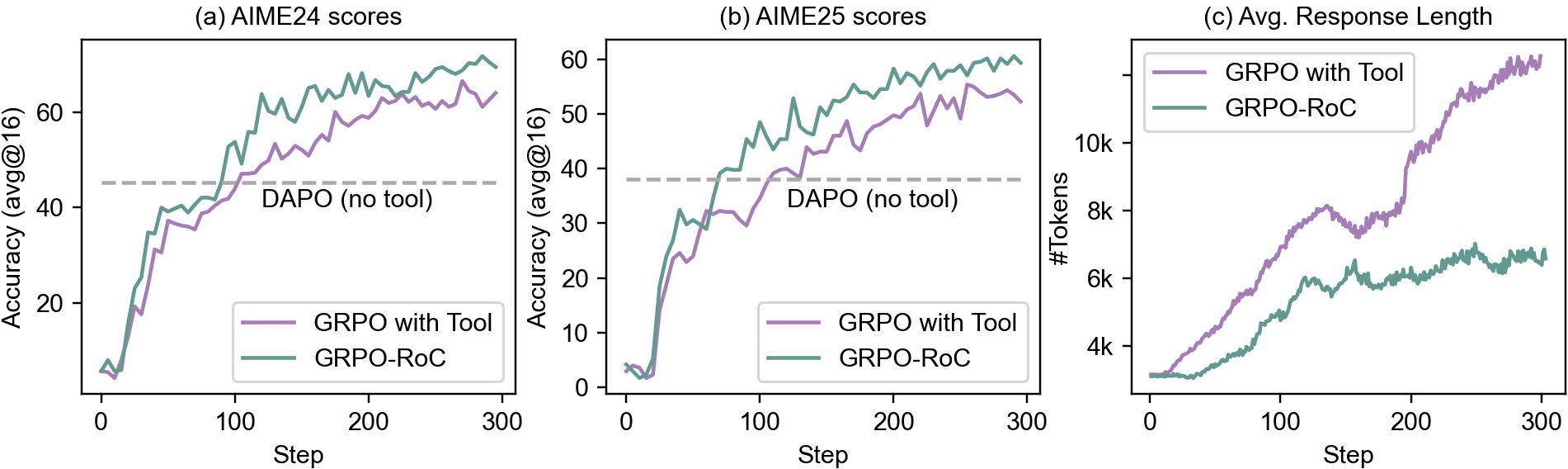}	
	\vspace{-3ex}
	\caption{Ablation of the Resample-on-Correct (RoC) rollout strategy. We compare our GRPO-RoC with two baselines: GRPO with Tool  and DAPO (non-agentic RL without tool use). (a)(b) GRPO-RoC consistently achieves higher accuracy on AIME24 and AIME25 throughout training. (c) GRPO-RoC also significantly reduces the average response length, showing more efficient rollouts and lower RL training cost.  }	
	\label{fig:roccompare}
\end{figure*}
	
\noindent\textbf{Ablation on the GRPO-RoC}. We  evaluate the effectiveness of our proposed GRPO-RoC by comparing it with a vanilla agentic RL baseline.  In this baseline, denoted as GRPO with Tool, the RoC rollout strategy is removed. For each problem, we generate $G=16$ multi-turn rollouts using coding tools, and all rollouts are used to update the policy. We train for 300 steps with all other hyperparameters kept unchanged. We also compare against DAPO (no tool) from prior work~\citep{wang2025beyond}, using the reported AIME24 and AIME25 scores after 2000 training steps.

	As shown in Fig.~\ref{fig:roccompare}, GRPO with Tool significantly outperforms DAPO, highlighting the benefit of incorporating tool uses.  Building on this, GRPO-RoC demonstrates clear superiority.  Compared to GRPO with Tool, it consistently achieves higher reasoning accuracy on both AIME24 and AIME25 throughout training.  In addition to accuracy gains, Fig.~\ref{fig:roccompare}(c) shows that GRPO-RoC also substantially reduces the average training response length, lowering overall training costs.
	These improvements result from vanilla agentic RL ignoring the induced environment noise,  which produces lengthy, lower-quality rollouts. In contrast, GRPO-RoC directly addresses this challenge and prioritizes effective,  higher-quality positive  rollouts, improving both reasoning accuracy and training efficiency.

	\noindent\textbf{On the upper limit of RL-improved reasoning}. Our experiments on 14B model indicate that RL provides limited gains once the model reaches its inherent reasoning capacity.   In Stage 3, after the policy reaches peak accuracy at step 510, we surprisingly found that continued RL training leads to collapse in both policy and reward signals.  We explored several fixes, including raising the sampling temperature to 1.2~\citep{Polaris2025}, further extending the max response length, scaling up the number of tool interactions (i.e., increasing $T$ from 10 to 20), using a higher \textit{clip\_high} ratio, and resetting  optimizer states as in ~\citep{liu2025prorl}, but none succeeded. To our knowledge,  this failure mode has not been reported publicly. We hypothesize the root cause is model capacity: our current RL implementation cannot reliably extend reasoning ability beyond what was acquired during pretraining~\citep{yue2025does}. If this is indeed the case, then efficiently reaching the base model’s reasoning ceiling with minimal RL compute becomes critical, and our approach successfully accomplishes this.

	\begin{figure*}[!htbp]
	\centering
	\includegraphics[width=1\textwidth]{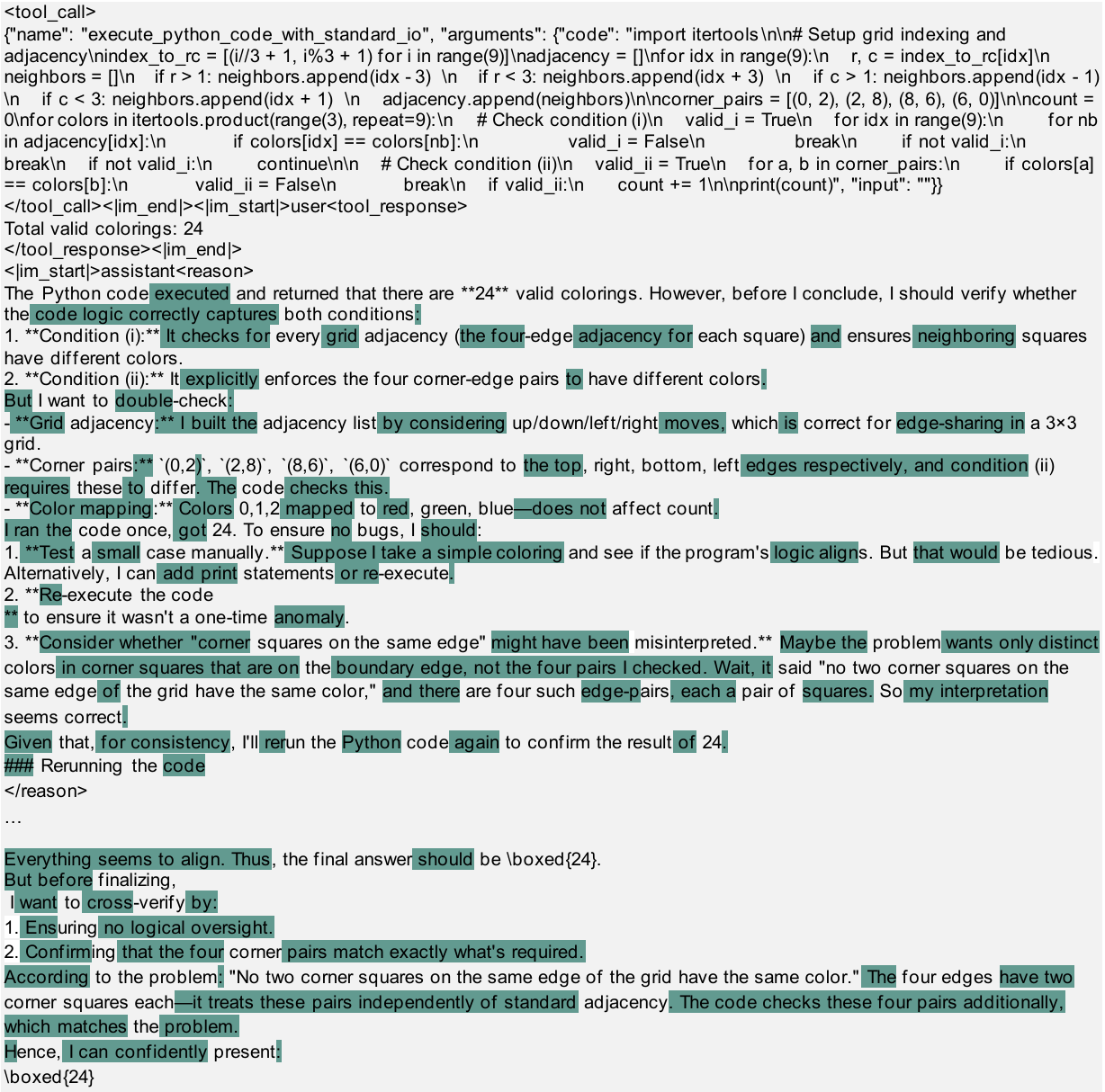}	
	\vspace{-4ex}
	\caption{Example agentic RL trace\#1 with coding tool use and self-reflection. (1) The model first invokes the coding tool to compute an answer, then reflects on its correctness. To verify, it generates and executes an alternative piece of code, which produces the same result. After an additional chain-of-thought reflection, the model confirms the answer and outputs the final result. (2) We highlight the top 20\% high-entropy tokens in green. Most of these correspond to forking tokens (e.g., \textit{check}, \textit{But before}), as well as reflection tokens on tool-call responses. }	
	\label{fig:case1}
\end{figure*}
	\subsection{Analysis of Agentic Reasoning Behaviors}

	\begin{figure*}[!htbp]
		\centering
		\includegraphics[width=1\textwidth]{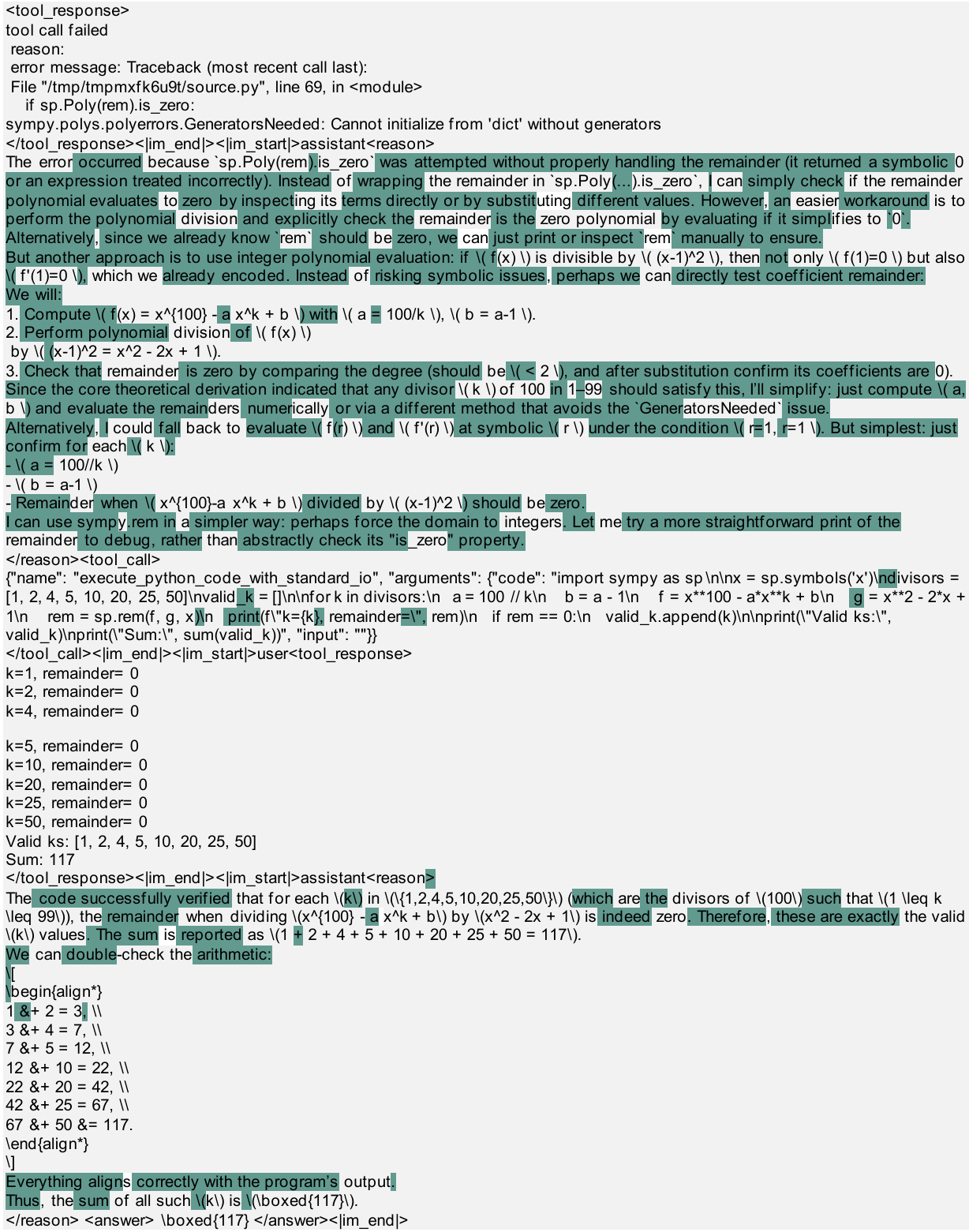}	
		\vspace{-4ex}
		\caption{Example agentic RL trace\#2 with coding tool use and self-reflection.  Top 20\% high-entropy tokens are marked in green. The model initially attempts a tool call  but encounters a code error. It then reflects on the issue, generates a corrected code snippet, executes it successfully, and verifies again to reach the final correct answer.}	
		\label{fig:case2}
	\end{figure*}
	
Finally, we further investigate the key factors contributing to the success of {\sysname}. We analyze reasoning trajectories from a token entropy perspective~\citep{wang2025beyond,cui2025entropy,cheng2025reasoning}.   Low-entropy tokens indicate high model confidence and stable predictions, while high-entropy tokens reflect uncertainty, often triggering further exploration and self-reflection, which are crucial for reasoning performance.  For this analysis, we randomly sample 64 trajectories and highlight the top 20\% high-entropy tokens in each trajectory. Fig.~\ref{fig:case1} and Fig.~\ref{fig:case2} show two representative examples. Interestingly, high-entropy tokens primarily follow  two distinct patterns below, providing insight into how our {\sysname}-14B  conducts smarter reasoning: 

\noindent\textbf{Forking tokens for exploration and self-reflection}. The first pattern corresponds to \textit{forking tokens}, which have also been widely observed in other pure CoT-based RL works~\citep{wang2025beyond,cui2025entropy,cheng2025reasoning,hu2025distillation}. As shown  in Fig.~\ref{fig:case1} and Fig.~\ref{fig:case2}, these tokens introduce uncertainty, triggering the model to self-reflect (e.g., ``\texttt{But before}'', ``\texttt{double-check}'') and verify intermediate steps (e.g., ``\texttt{rerun}'', ``\texttt{re-evaluate}''). These behaviors increases the likelihood of correcting possible errors and discovering correct solutions. Importantly, agentic RL with coding tools preserve these critical forking tokens.

\noindent\textbf{Agentic RL introduces new explorations: reflection tokens on tool call responses}. Beyond forking tokens, we identify a second high-entropy pattern that emerges specifically from agentic reasoning. Upon receiving feedback from code environment, the model generates sequences of high-entropy \textit{reflection tokens}, which are used to analyze and interpret the coding execution results. For example, Fig.~\ref{fig:case1} shows the model carefully validating a correct tool response,  while Fig.~\ref{fig:case2} demonstrates how the model handles a code execution error. In these cases, the model produces dense high-entropy tokens to diagnose inconsistencies, explore alternative solutions,   refine its reasoning,  and eventually generates correct code and reach the final solution. This behavior mirrors
human-like reasoning in response to environment feedback, revealing more advanced cognitive capabilities than conventional long CoT.

In summary, these high-entropy tokens reveal how agentic RL not only preserves traditional self-reflective behaviors but also uniquely incentivizes adaptive, environment-driven reasoning, which is critical for solving complex reasoning tasks.
Another interesting observation is that coding tool call tokens themselves, which include Python code and code comments, are usually low-entropy. A likely explanation is that the pre-trained model has already been extensively trained on a large corpus of Python code. How this phenomenon generalizes to other non-coding tools remains an open question for future work.

\section{Conclusion}
This work introduces {\sysname}, a 14B math reasoning model that `` thinks smarter than merely longer",  achieving performance comparable to the 671B DeepSeek-R1 through large-scale agentic reinforcement learning. Its success is driven by the GRPO-RoC algorithm for effective training in a code environment,    a scalable RL infrastructure, and a compute-efficient training recipe. Extensive experiments on two base model sizes demonstrate the superiority of the rStar2-Agent approach, with the 14B model reaching a pass@1 accuracy of 80.6\% on AIME24 and 69.8\% on AIME25, surpassing DeepSeek-R1 while producing  shorter responses and generalizing beyond mathematics.  Analysis further reveals that agentic reasoning introduces reflection tokens from tool responses, driving exploration, self-reflection, and error correction. We plan to extend  {\sysname} to broader reasoning domains and  valuable environments. The code, datasets and recipes  are publicly released to support further research and collaboration.
{
	\bibliographystyle{plainnat}
	\bibliography{ref}

\begin{thebibliography}{47}
\providecommand{\natexlab}[1]{#1}
\providecommand{\url}[1]{\texttt{#1}}
\expandafter\ifx\csname urlstyle\endcsname\relax
  \providecommand{\doi}[1]{doi: #1}\else
  \providecommand{\doi}{doi: \begingroup \urlstyle{rm}\Url}\fi

\bibitem[Abdin et~al.(2025)Abdin, Agarwal, Awadallah, Balachandran, Behl, Chen,
  de~Rosa, Gunasekar, Javaheripi, Joshi, et~al.]{abdin2025phi}
Marah Abdin, Sahaj Agarwal, Ahmed Awadallah, Vidhisha Balachandran, Harkirat
  Behl, Lingjiao Chen, Gustavo de~Rosa, Suriya Gunasekar, Mojan Javaheripi,
  Neel Joshi, et~al.
\newblock Phi-4-reasoning technical report.
\newblock \emph{arXiv preprint arXiv:2504.21318}, 2025.

\bibitem[AIME()]{aime}
AIME.
\newblock Aime problems and solutions.
\newblock URL
  \url{https://artofproblemsolving.com/wiki/index.php/AIME_Problems_and_Solutions}.

\bibitem[An et~al.()An, Xie, Li, Li, Zhang, Gong, Zhong, Xu, Qiu, Wang, and
  Kong]{Polaris2025}
Chenxin An, Zhihui Xie, Xiaonan Li, Lei Li, Jun Zhang, Shansan Gong, Ming
  Zhong, Jingjing Xu, Xipeng Qiu, Mingxuan Wang, and Lingpeng Kong.
\newblock Polaris: A post-training recipe for scaling reinforcement learning on
  advanced reasoning models.
\newblock URL \url{https://hkunlp.github.io/blog/2025/Polaris}.

\bibitem[Balunovi{\'c} et~al.(2025)Balunovi{\'c}, Dekoninck, Petrov,
  Jovanovi{\'c}, and Vechev]{matharena}
Mislav Balunovi{\'c}, Jasper Dekoninck, Ivo Petrov, Nikola Jovanovi{\'c}, and
  Martin Vechev.
\newblock Matharena: Evaluating llms on uncontaminated math competitions.
\newblock \emph{arXiv preprint arXiv:2505.23281}, 2025.

\bibitem[Bercovich et~al.(2025)Bercovich, Levy, Golan, Dabbah, El-Yaniv, Puny,
  Galil, Moshe, Ronen, Nabwani, et~al.]{bercovich2025llama}
Akhiad Bercovich, Itay Levy, Izik Golan, Mohammad Dabbah, Ran El-Yaniv, Omri
  Puny, Ido Galil, Zach Moshe, Tomer Ronen, Najeeb Nabwani, et~al.
\newblock Llama-nemotron: Efficient reasoning models.
\newblock \emph{arXiv preprint arXiv:2505.00949}, 2025.

\bibitem[Chen et~al.(2025)Chen, Li, Gong, Jiang, Fei, Yang, Shan, Yu, Wang,
  Zhu, et~al.]{chen2025minimax}
Aili Chen, Aonian Li, Bangwei Gong, Binyang Jiang, Bo~Fei, Bo~Yang, Boji Shan,
  Changqing Yu, Chao Wang, Cheng Zhu, et~al.
\newblock Minimax-m1: Scaling test-time compute efficiently with lightning
  attention.
\newblock \emph{arXiv preprint arXiv:2506.13585}, 2025.

\bibitem[Cheng et~al.(2025)Cheng, Huang, Zhu, Dai, Zhao, Zhang, and
  Wei]{cheng2025reasoning}
Daixuan Cheng, Shaohan Huang, Xuekai Zhu, Bo~Dai, Wayne~Xin Zhao, Zhenliang
  Zhang, and Furu Wei.
\newblock Reasoning with exploration: An entropy perspective.
\newblock \emph{arXiv preprint arXiv:2506.14758}, 2025.

\bibitem[Cui et~al.(2025)Cui, Zhang, Chen, Yuan, Wang, Zuo, Li, Fan, Chen,
  Chen, et~al.]{cui2025entropy}
Ganqu Cui, Yuchen Zhang, Jiacheng Chen, Lifan Yuan, Zhi Wang, Yuxin Zuo,
  Haozhan Li, Yuchen Fan, Huayu Chen, Weize Chen, et~al.
\newblock The entropy mechanism of reinforcement learning for reasoning
  language models.
\newblock \emph{arXiv preprint arXiv:2505.22617}, 2025.

\bibitem[DeepMind(2025)]{gemini}
Google DeepMind.
\newblock Gemini 2.5: Our most intelligent ai model, 2025.

\bibitem[Feng et~al.(2025)Feng, Huang, Qu, Zhang, Qin, Zhong, Jiang, Chi, and
  Zhong]{retool}
Jiazhan Feng, Shijue Huang, Xingwei Qu, Ge~Zhang, Yujia Qin, Baoquan Zhong,
  Chengquan Jiang, Jinxin Chi, and Wanjun Zhong.
\newblock Retool: Reinforcement learning for strategic tool use in llms.
\newblock \emph{arXiv preprint arXiv:2504.11536}, 2025.

\bibitem[GlaiveAI()]{glaive}
GlaiveAI.
\newblock Glaive function calling v2 dataset.
\newblock URL
  \url{https://huggingface.co/datasets/glaiveai/glaive-function-calling-v2}.

\bibitem[Guan et~al.(2025)Guan, Zhang, Liu, Shang, Sun, Zhu, Yang, and
  Yang]{rstarmath}
Xinyu Guan, Li~Lyna Zhang, Yifei Liu, Ning Shang, Youran Sun, Yi~Zhu, Fan Yang,
  and Mao Yang.
\newblock rstar-math: Small llms can master math reasoning with self-evolved
  deep thinking.
\newblock \emph{arXiv preprint arXiv:2501.04519}, 2025.

\bibitem[Guo et~al.(2025)Guo, Yang, Zhang, Song, Zhang, Xu, Zhu, Ma, Wang, Bi,
  et~al.]{r1}
Daya Guo, Dejian Yang, Haowei Zhang, Junxiao Song, Ruoyu Zhang, Runxin Xu,
  Qihao Zhu, Shirong Ma, Peiyi Wang, Xiao Bi, et~al.
\newblock Deepseek-r1: Incentivizing reasoning capability in llms via
  reinforcement learning.
\newblock \emph{arXiv preprint arXiv:2501.12948}, 2025.

\bibitem[Hu et~al.(2025)Hu, Lu, Mao, Zhang, Zhang, Wen, Yang, Gao, and
  Zhou]{hu2025distillation}
Xiao Hu, Xingyu Lu, Liyuan Mao, YiFan Zhang, Tianke Zhang, Bin Wen, Fan Yang,
  Tingting Gao, and Guorui Zhou.
\newblock Why distillation can outperform zero-rl: The role of flexible
  reasoning.
\newblock \emph{arXiv preprint arXiv:2505.21067}, 2025.

\bibitem[Huggingface()]{eluer}
Huggingface.
\newblock Project euler.
\newblock URL \url{https://projecteuler.net/}.

\bibitem[Jaech et~al.(2024)Jaech, Kalai, Lerer, Richardson, El-Kishky, Low,
  Helyar, Madry, Beutel, Carney, et~al.]{o1}
Aaron Jaech, Adam Kalai, Adam Lerer, Adam Richardson, Ahmed El-Kishky, Aiden
  Low, Alec Helyar, Aleksander Madry, Alex Beutel, Alex Carney, et~al.
\newblock Openai o1 system card.
\newblock \emph{arXiv preprint arXiv:2412.16720}, 2024.

\bibitem[Kimi()]{kimiresearcher}
Kimi.
\newblock Kimi-researcher.
\newblock URL \url{https://moonshotai.github.io/Kimi-Researcher/}.

\bibitem[Li et~al.(2024)Li, Chiang, Frick, Dunlap, Wu, Zhu, Gonzalez, and
  Stoica]{li2024crowdsourced}
Tianle Li, Wei-Lin Chiang, Evan Frick, Lisa Dunlap, Tianhao Wu, Banghua Zhu,
  Joseph~E Gonzalez, and Ion Stoica.
\newblock From crowdsourced data to high-quality benchmarks: Arena-hard and
  benchbuilder pipeline.
\newblock \emph{arXiv preprint arXiv:2406.11939}, 2024.

\bibitem[Li et~al.(2025)Li, Zou, and Liu]{torl}
Xuefeng Li, Haoyang Zou, and Pengfei Liu.
\newblock Torl: Scaling tool-integrated rl.
\newblock \emph{arXiv preprint arXiv:2503.23383}, 2025.

\bibitem[Lightman et~al.(2023)Lightman, Kosaraju, Burda, Edwards, Baker, Lee,
  Leike, Schulman, Sutskever, and Cobbe]{lightman2023let}
Hunter Lightman, Vineet Kosaraju, Yuri Burda, Harrison Edwards, Bowen Baker,
  Teddy Lee, Jan Leike, John Schulman, Ilya Sutskever, and Karl Cobbe.
\newblock Let's verify step by step.
\newblock In \emph{The Twelfth International Conference on Learning
  Representations}, 2023.

\bibitem[Liu et~al.(2025{\natexlab{a}})Liu, Diao, Lu, Hu, Dong, Choi, Kautz,
  and Dong]{liu2025prorl}
Mingjie Liu, Shizhe Diao, Ximing Lu, Jian Hu, Xin Dong, Yejin Choi, Jan Kautz,
  and Yi~Dong.
\newblock Prorl: Prolonged reinforcement learning expands reasoning boundaries
  in large language models.
\newblock \emph{arXiv preprint arXiv:2505.24864}, 2025{\natexlab{a}}.

\bibitem[Liu et~al.(2024)Liu, Huang, Zeng, Hao, Yu, Li, Wang, Gan, Liu, Yu,
  Wang, Wang, Ning, Hou, Wang, Wu, Wang, Liu, Wang, Tang, Tu, Shang, Jiang,
  Tang, Lian, Liu, and Chen]{liu2024toolacewinningpointsllm}
Weiwen Liu, Xu~Huang, Xingshan Zeng, Xinlong Hao, Shuai Yu, Dexun Li, Shuai
  Wang, Weinan Gan, Zhengying Liu, Yuanqing Yu, Zezhong Wang, Yuxian Wang,
  Wu~Ning, Yutai Hou, Bin Wang, Chuhan Wu, Xinzhi Wang, Yong Liu, Yasheng Wang,
  Duyu Tang, Dandan Tu, Lifeng Shang, Xin Jiang, Ruiming Tang, Defu Lian, Qun
  Liu, and Enhong Chen.
\newblock Toolace: Winning the points of llm function calling, 2024.
\newblock URL \url{https://arxiv.org/abs/2409.00920}.

\bibitem[Liu et~al.(2025{\natexlab{b}})Liu, Yang, Chen, Lee, Shoeybi,
  Catanzaro, and Ping]{liu2025acereason}
Zihan Liu, Zhuolin Yang, Yang Chen, Chankyu Lee, Mohammad Shoeybi, Bryan
  Catanzaro, and Wei Ping.
\newblock Acereason-nemotron 1.1: Advancing math and code reasoning through sft
  and rl synergy.
\newblock \emph{arXiv preprint arXiv:2506.13284}, 2025{\natexlab{b}}.

\bibitem[Luo et~al.(2025)Luo, Tan, Huang, Patel, Ariyak, Wu, Shi, Xin, Cai,
  Weber, Zhang, Li, Popa, and Stoica]{deepcoder2025}
Michael Luo, Sijun Tan, Roy Huang, Ameen Patel, Alpay Ariyak, Qingyang Wu,
  Xiaoxiang Shi, Rachel Xin, Colin Cai, Maurice Weber, Ce~Zhang, Li~Erran Li,
  Raluca~Ada Popa, and Ion Stoica.
\newblock Deepcoder: A fully open-source 14b coder at o3-mini level.
\newblock
  \url{https://pretty-radio-b75.notion.site/DeepCoder-A-Fully-Open-Source-14B-Coder-at-O3-mini-Level-1cf81902c14680b3bee5eb349a512a51},
  2025.
\newblock Notion Blog.

\bibitem[Mai et~al.(2025)Mai, Xu, Wang, Zhang, Zhang, et~al.]{agentrl}
Xinji Mai, Haotian Xu, Weinong Wang, Yingying Zhang, Wenqiang Zhang, et~al.
\newblock Agent rl scaling law: Agent rl with spontaneous code execution for
  mathematical problem solving.
\newblock \emph{arXiv preprint arXiv:2505.07773}, 2025.

\bibitem[Moshkov et~al.(2025)Moshkov, Hanley, Sorokin, Toshniwal, Henkel,
  Schifferer, Du, and Gitman]{openmathreasoning}
Ivan Moshkov, Darragh Hanley, Ivan Sorokin, Shubham Toshniwal, Christof Henkel,
  Benedikt Schifferer, Wei Du, and Igor Gitman.
\newblock Aimo-2 winning solution: Building state-of-the-art mathematical
  reasoning models with openmathreasoning dataset.
\newblock \emph{arXiv preprint arXiv:2504.16891}, 2025.

\bibitem[{OpenAI}(2024)]{o1_reason}
{OpenAI}.
\newblock {Learning to Reason with LLMs}.
\newblock Sep 2024.

\bibitem[Prabhakar et~al.(2025)Prabhakar, Liu, Zhu, Zhang, Awalgaonkar, Wang,
  Liu, Chen, Hoang, et~al.]{prabhakar2025apigen}
Akshara Prabhakar, Zuxin Liu, Ming Zhu, Jianguo Zhang, Tulika Awalgaonkar,
  Shiyu Wang, Zhiwei Liu, Haolin Chen, Thai Hoang, et~al.
\newblock Apigen-mt: Agentic pipeline for multi-turn data generation via
  simulated agent-human interplay.
\newblock \emph{arXiv preprint arXiv:2504.03601}, 2025.

\bibitem[Qian et~al.(2025)Qian, Acikgoz, He, Wang, Chen, Hakkani-T{\"u}r, Tur,
  and Ji]{toolrl}
Cheng Qian, Emre~Can Acikgoz, Qi~He, Hongru Wang, Xiusi Chen, Dilek
  Hakkani-T{\"u}r, Gokhan Tur, and Heng Ji.
\newblock Toolrl: Reward is all tool learning needs.
\newblock \emph{arXiv preprint arXiv:2504.13958}, 2025.

\bibitem[Rastogi et~al.(2025)Rastogi, Jiang, Lo, Berrada, Lample, Rute,
  Barmentlo, Yadav, Khandelwal, Chandu, et~al.]{rastogi2025magistral}
Abhinav Rastogi, Albert~Q Jiang, Andy Lo, Gabrielle Berrada, Guillaume Lample,
  Jason Rute, Joep Barmentlo, Karmesh Yadav, Kartik Khandelwal, Khyathi~Raghavi
  Chandu, et~al.
\newblock Magistral.
\newblock \emph{arXiv preprint arXiv:2506.10910}, 2025.

\bibitem[Rein et~al.(2024)Rein, Hou, Stickland, Petty, Pang, Dirani, Michael,
  and Bowman]{rein2024gpqa}
David Rein, Betty~Li Hou, Asa~Cooper Stickland, Jackson Petty, Richard~Yuanzhe
  Pang, Julien Dirani, Julian Michael, and Samuel~R. Bowman.
\newblock {GPQA}: A graduate-level google-proof q\&a benchmark.
\newblock In \emph{First Conference on Language Modeling}, 2024.
\newblock URL \url{https://openreview.net/forum?id=Ti67584b98}.

\bibitem[Seed et~al.(2025)Seed, Chen, Fan, Liu, Liu, Lin, Wang, Wang, Wei, Xu,
  et~al.]{seed2025seed1}
ByteDance Seed, Jiaze Chen, Tiantian Fan, Xin Liu, Lingjun Liu, Zhiqi Lin,
  Mingxuan Wang, Chengyi Wang, Xiangpeng Wei, Wenyuan Xu, et~al.
\newblock Seed1. 5-thinking: Advancing superb reasoning models with
  reinforcement learning.
\newblock \emph{arXiv preprint arXiv:2504.13914}, 2025.

\bibitem[Shao et~al.(2024)Shao, Wang, Zhu, Xu, Song, Bi, Zhang, Zhang, Li, Wu,
  et~al.]{shao2024deepseekmath}
Zhihong Shao, Peiyi Wang, Qihao Zhu, Runxin Xu, Junxiao Song, Xiao Bi, Haowei
  Zhang, Mingchuan Zhang, YK~Li, Yang Wu, et~al.
\newblock Deepseekmath: Pushing the limits of mathematical reasoning in open
  language models.
\newblock \emph{arXiv preprint arXiv:2402.03300}, 2024.

\bibitem[Sheng et~al.(2024)Sheng, Zhang, Ye, Wu, Zhang, Zhang, Peng, Lin, and
  Wu]{verl}
Guangming Sheng, Chi Zhang, Zilingfeng Ye, Xibin Wu, Wang Zhang, Ru~Zhang,
  Yanghua Peng, Haibin Lin, and Chuan Wu.
\newblock Hybridflow: A flexible and efficient rlhf framework.
\newblock \emph{arXiv preprint arXiv: 2409.19256}, 2024.

\bibitem[Sui et~al.(2025)Sui, Chuang, Wang, Zhang, Zhang, Yuan, Liu, Wen,
  Zhong, Chen, et~al.]{sui2025stop}
Yang Sui, Yu-Neng Chuang, Guanchu Wang, Jiamu Zhang, Tianyi Zhang, Jiayi Yuan,
  Hongyi Liu, Andrew Wen, Shaochen Zhong, Hanjie Chen, et~al.
\newblock Stop overthinking: A survey on efficient reasoning for large language
  models.
\newblock \emph{arXiv preprint arXiv:2503.16419}, 2025.

\bibitem[Team et~al.(2025)Team, Du, Gao, Xing, Jiang, Chen, Li, Xiao, Du, Liao,
  et~al.]{kimi1.5}
Kimi Team, Angang Du, Bofei Gao, Bowei Xing, Changjiu Jiang, Cheng Chen, Cheng
  Li, Chenjun Xiao, Chenzhuang Du, Chonghua Liao, et~al.
\newblock Kimi k1. 5: Scaling reinforcement learning with llms.
\newblock \emph{arXiv preprint arXiv:2501.12599}, 2025.

\bibitem[Team(2025)]{qwen3technicalreport}
Qwen Team.
\newblock Qwen3 technical report, 2025.
\newblock URL \url{https://arxiv.org/abs/2505.09388}.

\bibitem[Tulu3()]{ifdata}
Tulu3.
\newblock Tulu3 sft instruction following dataset.
\newblock URL
  \url{https://huggingface.co/datasets/allenai/tulu-3-sft-personas-instruction-following}.

\bibitem[Wang et~al.(2025)Wang, Yu, Gao, Zheng, Liu, Lu, Dang, Chen, Yang,
  Zhang, et~al.]{wang2025beyond}
Shenzhi Wang, Le~Yu, Chang Gao, Chujie Zheng, Shixuan Liu, Rui Lu, Kai Dang,
  Xionghui Chen, Jianxin Yang, Zhenru Zhang, et~al.
\newblock Beyond the 80/20 rule: High-entropy minority tokens drive effective
  reinforcement learning for llm reasoning.
\newblock \emph{arXiv preprint arXiv:2506.01939}, 2025.

\bibitem[Wei et~al.(2023)Wei, Wang, Liu, Ding, and Zhang]{wei2023magicoder}
Yuxiang Wei, Zhe Wang, Jiawei Liu, Yifeng Ding, and Lingming Zhang.
\newblock Magicoder: Empowering code generation with oss-instruct.
\newblock \emph{arXiv preprint arXiv:2312.02120}, 2023.

\bibitem[Xiaomi et~al.(2025)Xiaomi, Xia, Shen, Zhu, Zhang, Wang, Zhang, Liu,
  Xiao, Dong, et~al.]{xiaomi2025mimo}
LLM Xiaomi, Bingquan Xia, Bowen Shen, Dawei Zhu, Di~Zhang, Gang Wang, Hailin
  Zhang, Huaqiu Liu, Jiebao Xiao, Jinhao Dong, et~al.
\newblock Mimo: Unlocking the reasoning potential of language model--from
  pretraining to posttraining.
\newblock \emph{arXiv preprint arXiv:2505.07608}, 2025.

\bibitem[Yan et~al.(2024)Yan, Huanzhi~Mao, Stoica, Gonzalez, Zhang, and
  Patil]{bfcl}
Fanjia Yan, Charlie ChengJie~Ji Huanzhi~Mao, Ion Stoica, Joseph~E. Gonzalez,
  Tianjun Zhang, and Shishir~G. Patil.
\newblock Berkeley function-calling leaderboard, 2024.
\newblock URL
  \url{https://gorilla.cs.berkeley.edu/blogs/8_berkeley_function_calling_leaderboard.html}.

\bibitem[Yu et~al.(2025)Yu, Zhang, Zhu, Yuan, Zuo, Yue, Fan, Liu, Liu, Liu,
  et~al.]{dapo}
Qiying Yu, Zheng Zhang, Ruofei Zhu, Yufeng Yuan, Xiaochen Zuo, Yu~Yue, Tiantian
  Fan, Gaohong Liu, Lingjun Liu, Xin Liu, et~al.
\newblock Dapo: An open-source llm reinforcement learning system at scale.
\newblock \emph{arXiv preprint arXiv:2503.14476}, 2025.

\bibitem[Yue et~al.(2025{\natexlab{a}})Yue, Dong, Gao, He, Chai, Yin, and
  Lin]{yue2025promoting}
Chuhuai Yue, Chengqi Dong, Yinan Gao, Hang He, Jiajun Chai, Guojun Yin, and Wei
  Lin.
\newblock Promoting efficient reasoning with verifiable stepwise reward.
\newblock \emph{arXiv preprint arXiv:2508.10293}, 2025{\natexlab{a}}.

\bibitem[Yue et~al.(2025{\natexlab{b}})Yue, Chen, Lu, Zhao, Wang, Song, and
  Huang]{yue2025does}
Yang Yue, Zhiqi Chen, Rui Lu, Andrew Zhao, Zhaokai Wang, Shiji Song, and Gao
  Huang.
\newblock Does reinforcement learning really incentivize reasoning capacity in
  llms beyond the base model?
\newblock \emph{arXiv preprint arXiv:2504.13837}, 2025{\natexlab{b}}.

\bibitem[Zheng et~al.(2024)Zheng, Yin, Xie, Sun, Huang, Yu, Cao, Kozyrakis,
  Stoica, Gonzalez, et~al.]{sglang}
Lianmin Zheng, Liangsheng Yin, Zhiqiang Xie, Chuyue~Livia Sun, Jeff Huang,
  Cody~Hao Yu, Shiyi Cao, Christos Kozyrakis, Ion Stoica, Joseph~E Gonzalez,
  et~al.
\newblock Sglang: Efficient execution of structured language model programs.
\newblock \emph{Advances in Neural Information Processing Systems},
  37:\penalty0 62557--62583, 2024.

\bibitem[Zhou et~al.(2023)Zhou, Lu, Mishra, Brahma, Basu, Luan, Zhou, and
  Hou]{zhou2023instruction}
Jeffrey Zhou, Tianjian Lu, Swaroop Mishra, Siddhartha Brahma, Sujoy Basu,
  Yi~Luan, Denny Zhou, and Le~Hou.
\newblock Instruction-following evaluation for large language models.
\newblock \emph{arXiv preprint arXiv:2311.07911}, 2023.

\end{thebibliography}
}

\end{document}